\pdfoutput=1
\documentclass{article}

\usepackage[T1]{fontenc}
\usepackage{textcomp}
\usepackage{microtype}
\usepackage{graphicx}
\usepackage{booktabs}
\usepackage{hyperref}

\PassOptionsToPackage{numbers,square,comma,sort&compress}{natbib}
\usepackage[preprint]{icml2027}
\setcitestyle{numbers,square,comma,sort&compress}
\makeatletter
\renewcommand{\ICML@preprint}{\textit{Preprint.}}
\makeatother
\hypersetup{
  pdftitle={Measure Then Revise: Budgeted Compute Routers From a Harness Self-Improvement Loop},
  pdfauthor={Tanmay Sah, Dolly Sah, Harshul Jain, Tanya Sah, Saroj Pradhan},
  pdfsubject={Preprint}
}
\usepackage{amsmath}
\usepackage{amssymb}
\usepackage{mathtools}
\usepackage{algorithm}
\usepackage{algorithmic}
\usepackage{enumitem}
\usepackage{placeins}
\usepackage[table]{xcolor}
\usepackage{listings}
\usepackage[most]{tcolorbox}
\usepackage[capitalize,noabbrev]{cleveref}

\definecolor{cardborder}{HTML}{E5E7EB}
\definecolor{codebg}{HTML}{F3F4F6}
\definecolor{codelabel}{HTML}{6B7280}
\definecolor{codekw}{HTML}{5B21B6}
\definecolor{codenum}{HTML}{9A3412}
\definecolor{codetext}{HTML}{1F2937}
\definecolor{diffaddbg}{HTML}{E6F7ED}
\definecolor{diffaddfg}{HTML}{065F46}
\definecolor{diffrembg}{HTML}{FDE8E8}
\definecolor{diffremfg}{HTML}{991B1B}
\definecolor{diffhunkbg}{HTML}{EEF2FF}
\definecolor{diffhunkfg}{HTML}{3730A3}
\definecolor{diffctxbg}{HTML}{F8FAFC}
\definecolor{diffctxfg}{HTML}{334155}

\newcommand{\diffline}[3]{%
  \noindent\colorbox{#1}{\parbox{\dimexpr\linewidth-2\fboxsep\relax}{\ttfamily\scriptsize\color{#2}\strut #3}}\par\nointerlineskip
}
\newcommand{\difffileold}[1]{\diffline{diffrembg}{diffremfg}{#1}}
\newcommand{\difffilenew}[1]{\diffline{diffaddbg}{diffaddfg}{#1}}
\newcommand{\diffhunk}[1]{\diffline{diffhunkbg}{diffhunkfg}{#1}}
\newcommand{\diffrem}[1]{\diffline{diffrembg}{diffremfg}{#1}}
\newcommand{\diffadd}[1]{\diffline{diffaddbg}{diffaddfg}{#1}}
\newcommand{\diffctx}[1]{\diffline{diffctxbg}{diffctxfg}{#1}}

\lstdefinestyle{bashcard}{
  language=bash,
  basicstyle=\ttfamily\footnotesize\color{codetext},
  keywordstyle=\color{codekw},
  morekeywords={export,python3},
  literate={0}{{{\color{codenum}0}}}1
           {1}{{{\color{codenum}1}}}1
           {2}{{{\color{codenum}2}}}1
           {3}{{{\color{codenum}3}}}1
           {4}{{{\color{codenum}4}}}1
           {5}{{{\color{codenum}5}}}1
           {6}{{{\color{codenum}6}}}1
           {8}{{{\color{codenum}8}}}1,
  columns=fullflexible,
  keepspaces=true,
  showstringspaces=false,
  upquote=true,
  aboveskip=0pt,
  belowskip=0pt,
  xleftmargin=2pt,
  xrightmargin=2pt
}

\newcommand{\codepill}[1]{\tcbox[on line, nobeforeafter, tcbox raise base, arc=2.5pt, colback=codebg, colframe=codebg, boxsep=0.6pt, left=2.5pt, right=2.5pt, top=1.0pt, bottom=1.0pt]{\ttfamily\footnotesize\color{codetext}#1}}

\newenvironment{listingcard}[1]{%
  \begin{tcolorbox}[enhanced, breakable=false, colback=white, colframe=cardborder, boxrule=0.6pt, arc=5pt, left=9pt, right=9pt, top=7pt, bottom=8pt, before skip=6pt, after skip=8pt]
  {\footnotesize\sffamily\color{codelabel}#1}\par\vspace{5pt}%
  \begin{tcolorbox}[enhanced, colback=codebg, colframe=codebg, boxrule=0pt, arc=4.5pt, left=9pt, right=9pt, top=7pt, bottom=7pt, before skip=0pt, after skip=0pt]
}{%
  \end{tcolorbox}%
  \end{tcolorbox}%
}

\newcommand{\pmsd}[2]{$#1 \pm #2$}

\icmltitlerunning{Measure Then Revise: Budgeted Compute Routers From a Harness Self-Improvement Loop}

\begin{document}

\twocolumn[
\icmltitle{Measure Then Revise: Budgeted Compute Routers From a \\ Harness Self-Improvement Loop}

\begin{icmlauthorlist}
\icmlauthor{Tanmay Sah}{ind}
\icmlauthor{Dolly Sah}{ind}
\icmlauthor{Harshul Jain}{ind}
\icmlauthor{Tanya Sah}{ind}
\icmlauthor{Saroj Pradhan}{gt}
\end{icmlauthorlist}

\icmlaffiliation{ind}{Independent Researcher}
\icmlaffiliation{gt}{Georgia Institute of Technology}
\icmlcorrespondingauthor{Tanmay Sah}{tradertanmay@gmail.com}
\icmlcorrespondingauthor{Dolly Sah}{dolly17sah@gmail.com}
\icmlcorrespondingauthor{Harshul Jain}{harshuljain1393@gmail.com}
\icmlcorrespondingauthor{Tanya Sah}{tanyasah20@gmail.com}
\icmlcorrespondingauthor{Saroj Pradhan}{spradhan75@gatech.edu}

\icmlkeywords{Agent Harness, Recursive Self-Improvement, Test-Time Compute, Evaluation Methodology, Tau-Bench}

\vskip 0.3in
]

\printAffiliationsAndNotice{}

\begin{abstract}
Automated harness self-improvement modifies the prompts and code surrounding a frozen language model. However, reported improvements often overlook generalization, inference costs, and evaluation noise. We investigate these limitations across 75 full-loop runs, 966 harnesses, and over 445,000 $\tau$-bench conversations using unseen tasks and a $2\times$ executor output-token budget. Prompt guidelines and execution guards produce little improvement on unseen tasks ($-3.0$ to $0.2$ percentage points). In contrast, Python-based compute routers that dynamically allocate reasoning effort generalize across five proposer models, with $95.6\%$ ($172/180$) outperforming their initial harness. However, $60.6\%$ ($109/180$) exceed the token budget. We introduce \emph{measure-then-revise}, a method that uses measured token consumption at each reasoning level to revise expensive routers rather than discard them. Across six GLM-5.3 seeds per method under search-only selection, budget violations decrease from $38/72$ to $2/72$, and wasted search episodes decrease by $53\%$. All six revised seeds achieve positive held-out gains, averaging \pmsd{11.6}{2.9} percentage points versus \pmsd{5.4}{6.4} under discard-and-resample. Comparisons with uniform-effort baselines and three reimplemented compute allocators show that automated harness evolution can reach performance comparable to uniform medium effort without weight training.

\vspace{4pt}
\noindent\textbf{Code Availability:} \url{https://github.com/tradertanmay/measure-then-revise}
\end{abstract}

\section{Introduction}
\label{sec:intro}

An agent harness is the set of prompts and code that controls how a language model operates without changing its weights. It manages system instructions, execution checks, retry logic, and inference settings. Recent studies have explored automating harness design by using a proposer model to analyze execution traces, modify the harness, and retain variants that perform better \citep{hu2024,zhang2024,robeyns2025,zhang2025,wang2025,lee2026b}.

\begin{figure*}[t]
\centering
\includegraphics[width=\textwidth]{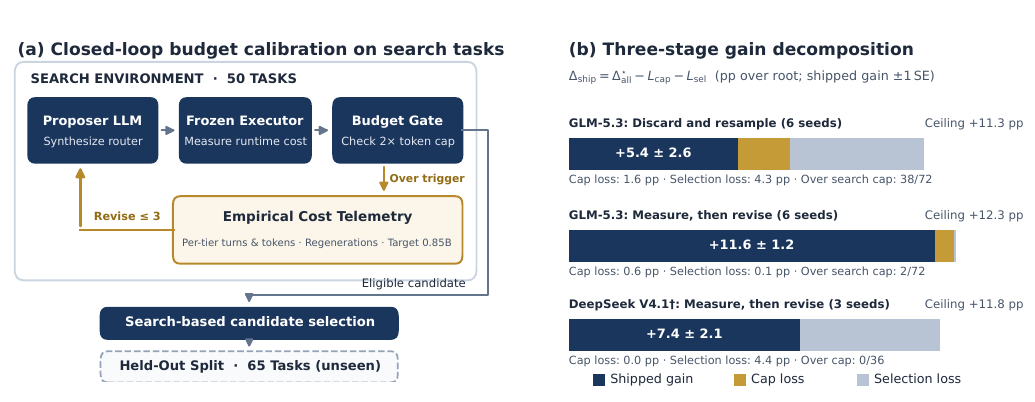}
\vspace{-10pt}
\caption{Closed-loop budget calibration and three-stage gain decomposition. (a) Search-split architecture showing runtime cost gating, per-tier telemetry feedback, candidate selection, and unseen held-out evaluation. (b) Additive decomposition $\Delta_{\mathrm{ship}} = \Delta_{\mathrm{all}}^\star - L_{\mathrm{cap}} - L_{\mathrm{sel}}$ under search-only eligibility on GLM-5.3 and combined-cost accounting on DeepSeek$^\dagger$.}
\label{fig:overview}
\vspace{-4pt}
\end{figure*}

Although these methods report substantial improvements, three limitations make their results difficult to interpret. First, candidate harnesses are often evaluated on the same tasks used to optimize them \citep{wang2026}. Second, higher accuracy can come at the cost of increased token consumption, which is not always measured \citep{kapoor2024,ding2026}. Third, only $22\%$ of recursive self-improvement studies audit multiple independent runs of the full optimization loop \citep{chen2026}. To address these limitations, we separate search tasks from held-out evaluation tasks and repeat our experiments across multiple independent seeds (\cref{fig:overview}). We also impose a token budget that limits the mean executor output tokens per episode to twice that of the initial harness.

We research whether automated harness evolution can improve agent performance by changing how reasoning effort is allocated rather than changing what the prompts say. Instead of assigning the same reasoning effort to every turn, a compute router can determine when additional reasoning is needed. Existing methods address this problem by training compute allocation policies \citep{yang2026ares,zhang2026baar,jali2026}. In contrast, we study whether a harness evolution loop can discover and refine such routing strategies through executable code, without weight training.

Using a fixed open-weight executor (\texttt{gpt-oss-120b}), task splits, and token budget on $\tau$-bench \citep{yao2024}, we make four contributions:
\begin{enumerate}[leftmargin=14pt,itemsep=2.5pt,parsep=0pt,topsep=3pt]
    \item \textbf{Measure-then-revise (\cref{sec:method_trim,sec:trim}).} Instead of discarding expensive routers, we provide the proposer with measured token consumption at each reasoning level to revise existing routers, reducing search-budget violations from $38/72$ to $2/72$ and wasted search episodes by $53\%$ while achieving held-out gains across all GLM-5.3 runs.
    \item \textbf{Two-stage compute routers vs.\ prompt and guard edits (\cref{sec:method,sec:prose,sec:routers}).} Whereas prompt instructions and pre-write guards cause regressions on previously successful tasks ($-3.0$ to $0.2$\,pp), two-stage Python routers that dynamically allocate reasoning effort before generation and before tool execution generalize across all five proposer models ($95.6\%$ outperform the initial harness).
    \item \textbf{Three-stage gain decomposition and the budget bottleneck (\cref{sec:setup,sec:routers}).} We separate candidate quality, budget-related losses, and candidate selection losses under held-out evaluation, showing that $60.6\%$ of open-loop routers exceed the $2\times$ token budget without runtime cost feedback.
    \item \textbf{Comparisons with uniform effort and learned allocators (\cref{sec:controls}).} Through uniform-effort baselines, three reimplemented compute allocators, and supplementary selection, domain-transfer, and executor analyses (Appendices~\ref{app:selection} and~\ref{app:partitions}), we show that harness evolution can autonomously reach performance comparable to uniform medium effort without weight training.
\end{enumerate}

\section{Related Work}
\label{sec:related}

\subsection{Self-Improving Agents and Harness Search}
Prior work has explored evolving agent code scaffolds \citep{zelikman2023,yin2024,robeyns2025,zhang2025,wang2025,zhang2026}, optimizing workflows around frozen language models \citep{hu2024,zhang2024,lee2026b,shang2026,jiang2026harnessmm,sengupta2026,hong2026,ling2026,zhang2026cig,wei2026}, jointly evolving model weights and harnesses \citep{luo2026}, and using programmatic scaffolds to keep individual model calls within their expected input distribution \citep{zhang2026compgen}. Other approaches use execution feedback to revise prompt instructions \citep{yang2023,guo2023,fernando2023,khattab2023,opsahlong2024,yuksekgonul2024,cheng2024,agrawal2025}.

Rather than introducing a new evolutionary loop, we research how the choice of harness modifications and computational constraints affects generalization with frozen models. Modifying prompt instructions can cause regressions on previously successful tasks, whereas per-turn effort routing adjusts reasoning compute without altering the prompt.

\subsection{Evaluation Pitfalls in Agent Optimization}
Previous studies identify several challenges in evaluating agent optimization, including task leakage and unmeasured increases in inference compute \citep{kapoor2024,chen2026,wang2026,ding2026,ding2026b}. Modifications intended to correct specific failures can also introduce new failures in previously successful tasks \citep{he2026}. Limited evaluation sets can lead to benchmark saturation and favor candidates that perform well by chance \citep{song2024,huang2024,tian2026,xie2026}.

We evaluate candidates using paired task-level comparisons \citep{miller2024} and introduce a three-stage gain decomposition that distinguishes the best observed candidate performance, the improvement lost due to budget constraints, and the improvement lost during candidate selection.

\subsection{Adaptive Test-Time Compute}
Increasing inference compute can improve reasoning performance \citep{brown2024,snell2024}, although excessive reasoning can sometimes reduce accuracy \citep{gema2025,lee2026}. Existing approaches train compute allocation policies through reinforcement learning, boundary supervision, or dual optimization \citep{yang2026ares,zhang2026baar,jali2026,zhai2026,kassenaar2026}. Other methods allocate compute based on model confidence, the consequences of errors, or resource constraints \citep{he2026b,sah2026,madiraju2026}.

Our work researches an alternative approach in which a proposer model generates executable compute routers within an automated harness evolution loop. Rather than training router weights, we use measured token consumption to revise routing code and improve budget compliance.

\section{Problem Setup and Evaluation Design}
\label{sec:setup}

\subsection{Budget-Constrained Harness Self-Improvement}
We consider an executor $M_\theta$ with frozen weights that interacts with a stochastic environment through a candidate harness $h \in \mathcal{H}$. Each episode $\tau$, executed on task $i$ with seed $s$, produces a binary reward $r(\tau) \in \{0, 1\}$ and an executor output token count $c(\tau) \in \mathbb{N}$.

Before running the experiments, we divide the tasks into two non-overlapping sets: a search split $\mathcal{S}$, used to develop and rank candidate harnesses, and a held-out split $\mathcal{T}$, used to evaluate generalization.

Evaluating a harness on $\mathcal{S}$ produces its search score $\hat{R}_{\mathcal{S}}(h)$ and mean token cost $\hat{C}_{\mathcal{S}}(h)$. Evaluating it on $\mathcal{T}$ produces its held-out score $R_{\mathcal{T}}(h)$. We denote the mean output token count across both splits by $C(h)$. Neither the proposer nor the candidate selection process has access to held-out rewards.

\begin{table}[t]
\caption{Experimental setup on $\tau$-bench retail.}
\label{tab:protocol}
\vspace{2pt}
\centering
\scriptsize
\setlength{\tabcolsep}{4.5pt}
\begin{tabular}{@{}ll@{}}
\toprule
Component & Configuration \\
\midrule
\multicolumn{2}{@{}l}{\textit{Executor \& Serving}} \\
Executor & \texttt{openai/gpt-oss-120b}, frozen weights (temp $0$) \\
Customer simulator & \texttt{openai/gpt-oss-120b} (temp $1.0$) \\
Serving stack & vLLM, $1\times$ NVIDIA H100 80\,GB GPU \\
\midrule
\multicolumn{2}{@{}l}{\textit{Evaluation Splits}} \\
Search split ($\mathcal{S}$) & $50$ retail tasks (seed 4321) \\
Held-out split ($\mathcal{T}$) & $65$ unseen retail tasks ($\mathcal{S} \cap \mathcal{T} = \emptyset$) \\
Rollouts per task & $3$ ($150$ search $+ 195$ held-out $= 345$ eps/harness) \\
\midrule
\multicolumn{2}{@{}l}{\textit{Search Loop}} \\
Search schedule & $4$ search rounds $\times 3$ candidates ($12$ new candidates) \\
Evaluated per seed & $13$ harnesses ($1$ root $+ 12$ candidates; $4{,}485$ eps) \\
Independent seeds & $3$ per setting; GLM-5.3 comparison uses $6$ \\
\midrule
\multicolumn{2}{@{}l}{\textit{Budget \& Calibration}} \\
Mean token budget ($B$) & $2.0\times$ root mean output tokens per episode \\
Fast-screen trigger & $100\%$ of budget ($1.00B$ after Rollout 1, $50$ eps) \\
Full-search trigger & $95\%$ of budget ($0.95B$ across $150$ search eps) \\
Revision target & $85\%$ of budget ($0.85B$), max $K = 3$ revisions \\
\bottomrule
\end{tabular}
\vspace{-6pt}
\end{table}

In the main experiments, each independent loop starts from a root harness $h_0$, which uses low reasoning effort and no additional instructions (\cref{tab:protocol}). The loop runs four search rounds, generating three candidates per round from the highest-scoring eligible harness found so far. Including the root, a complete loop evaluates 13 harnesses:
\[
\mathcal{A} = \{h_0, \ldots, h_{12}\}.
\]

We impose a token budget of twice the root harness's mean output token count ($\beta = 2.0$). All online checks for discarding or revising candidates use only search episodes.

We define two budget eligibility rules. Our primary rule, used in our six-seed comparisons, determines eligibility exclusively from search episodes:
\[
\mathcal{E}_{\mathcal{S}} = \{h \in \mathcal{A} : \hat{C}_{\mathcal{S}}(h) \le 2\hat{C}_{\mathcal{S}}(h_0)\},
\]
and selects the final harness without access to held-out rewards or held-out token counts:
\[
h^\star = \arg\max_{h \in \mathcal{E}_{\mathcal{S}}} \hat{R}_{\mathcal{S}}(h).
\]
For our three-seed multi-proposer sweeps and retrospective comparisons, we also evaluate combined-cost eligibility across both search and held-out episodes:
\[
\mathcal{E} = \{h \in \mathcal{A} : C(h) \le 2C(h_0)\},
\]
replacing $\mathcal{E}_{\mathcal{S}}$ with $\mathcal{E}$ during archive selection (accounting details in Appendix~\ref{app:ledger}).

\subsection{Three-Stage Gain Decomposition}
A harness self-improvement loop may fail to achieve its full potential for three reasons: the proposer generates weak candidates, strong candidates exceed the token budget, or the selection process fails to identify the best eligible candidate.

To distinguish these factors, we define two oracle gains over the evaluated archive, where $\mathcal{E}' \in \{\mathcal{E}_{\mathcal{S}}, \mathcal{E}\}$ and $h_0 \in \mathcal{E}' \subseteq \mathcal{A}$.

The unconstrained oracle measures the largest held-out improvement among all evaluated harnesses, regardless of token consumption:
\[
\Delta^\star_{\mathrm{all}} \coloneqq \max_{h \in \mathcal{A}} \left(R_{\mathcal{T}}(h) - R_{\mathcal{T}}(h_0)\right).
\]
The eligible oracle measures the largest held-out improvement among harnesses that satisfy the token budget:
\[
\Delta^\star_{\mathrm{elig}} \coloneqq \max_{h \in \mathcal{E}'} \left(R_{\mathcal{T}}(h) - R_{\mathcal{T}}(h_0)\right).
\]
The shipped gain is the held-out improvement achieved by the selected harness:
\[
\Delta_{\mathrm{ship}} \coloneqq R_{\mathcal{T}}(h^\star) - R_{\mathcal{T}}(h_0).
\]

Because $h^\star \in \mathcal{E}' \subseteq \mathcal{A}$, we decompose the shipped gain as:
\begin{equation}
\label{eq:decomp}
\Delta_{\mathrm{ship}}
=
\underbrace{\Delta^\star_{\mathrm{all}}}_{\text{Ceiling}}
-
\underbrace{(\Delta^\star_{\mathrm{all}} - \Delta^\star_{\mathrm{elig}})}_{\text{Cap loss } L_{\mathrm{cap}} \ge 0}
-
\underbrace{(\Delta^\star_{\mathrm{elig}} - \Delta_{\mathrm{ship}})}_{\text{Selection loss } L_{\mathrm{sel}} \ge 0}.
\end{equation}

Cap loss measures the potential improvement lost because candidates exceed the token budget. Selection loss measures the difference between the best eligible harness and the harness selected using search performance.

Both oracles represent the best observed performance among the evaluated candidates, rather than theoretical upper bounds. Including the root harness ensures that both oracle gains are non-negative. In negative controls where all generated candidates perform worse than the root, we also report the candidate-only maximum gain, $\Delta^\star_{\mathrm{cand}}$, to quantify this regression.

Finally, we report two additional metrics: $\rho_{\mathrm{start}}$, the proportion of generated candidates that outperform the root on held-out tasks, and $\bar{\tau}_{\mathrm{step}}$, the mean held-out performance difference between candidates and their respective parent harnesses.

\subsection{Experimental Configuration and Outcome Variance}
Multi-turn customer simulations introduce variability in task outcomes, even when the harness remains unchanged. Across 12 runs of the identical all-\texttt{low} root harness on $\tau$-bench retail \citep{yao2024}, $63\%$ of tasks produce both successful and unsuccessful outcomes across rollouts.

Differences in task difficulty account for $69\%$ of episode-level variance, while rollout noise accounts for the remaining $31\%$. The held-out standard error for a single seed is approximately $3.5$ percentage points.

To account for this variability, we report the mean and standard deviation of performance across independent loop seeds, together with task-paired $t$-statistics ($t_{\mathrm{task}}$) computed over the $n = 65$ held-out retail tasks ($n = 30$ for airline domain transfer and $n = 115$ for train-to-test settings; Appendix~\ref{app:ledger}). \Cref{tab:protocol} summarizes the experimental configuration.

\section{Edit Spaces and Measure-Then-Revise}
\label{sec:method}

\subsection{From Textual Guidelines to Two-Stage Compute Routers}
Existing harness improvement methods often modify system prompts or introduce checks before executing write operations. We evaluate three types of modifications: concise guidelines (up to 150 words), extended guidelines (up to 400 words), and Python-based pre-write guards implemented through \texttt{check(tool, args, ctx)}. Depending on the experimental setting, the proposer receives failed execution traces, differences in write actions, or pairs of successful and unsuccessful trajectories.

\begin{figure}[t]
\centering
\includegraphics[width=\columnwidth]{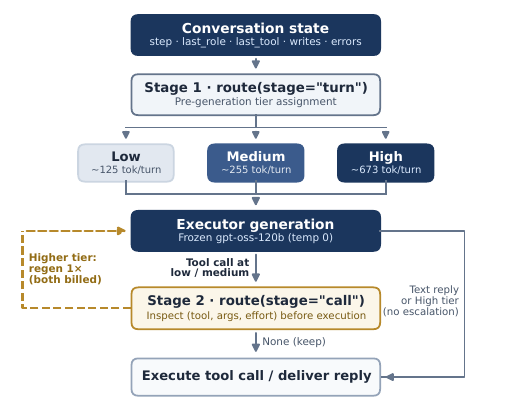}
\vspace{-10pt}
\caption{Two-stage compute router execution flow. Stage 1 (\texttt{stage="turn"}) assigns pre-generation reasoning effort; Stage 2 (\texttt{stage="call"}) inspects proposed tool calls from \texttt{low}/\texttt{medium} turns and optionally escalates once before execution (full rules in \cref{tab:interface}).}
\label{fig:router_exec}
\vspace{-6pt}
\end{figure}

Instead of modifying prompt instructions, we introduce a Python-based compute router, \texttt{route(stage, info, ctx)}, that controls the reasoning effort of the executor at two stages (\cref{fig:router_exec}).

\textbf{Turn stage.} Before the executor generates a response, the router examines the current conversation state, including the step count, recent messages, previous tool calls, completed write operations, and encountered errors. Based on this information, it assigns low, medium, or high reasoning effort.

\textbf{Call stage.} When a response generated at low or medium effort contains a tool call, the router examines the proposed tool name and arguments. If additional reasoning is considered necessary, the router can regenerate the response once at a higher effort level before executing the tool call. Output tokens from both the original and regenerated responses count toward the episode's token consumption.

The router therefore controls how much reasoning effort the executor uses without directly modifying prompt instructions or executing tools.

All routers are limited to 150 lines of Python code and are checked for valid Python syntax and function signature before evaluation. They cannot directly access environment state, tool implementations, domain policies, or evaluation results. The complete router interface and validation rules are provided in Appendix~\ref{app:prompts}.

\begin{algorithm}[!t]
\caption{Two-Stage Measure-Then-Revise Gate}
\label{alg:trim}
\footnotesize
\begin{algorithmic}[1]
\STATE {\bfseries Input:} Draft $h$, cap $B{=}2\hat{C}_{\mathcal{S}}(h_0)$, search split $\mathcal{S}$, seeds $(s_1,s_2,s_3)$, proposer $P$, max revisions $K{=}3$, $\gamma_{\mathrm{trig}}{=}0.95$, $\gamma_{\mathrm{targ}}{=}0.85$
\FOR{$k = 0, \dots, K$}
    \STATE $E_1 \leftarrow \textsc{Eval}(h, \mathcal{S}, \{s_1\})$; \; $\hat{c}_1 \leftarrow \textsc{MeanOutTok}(E_1)$
    \IF{$\hat{c}_1 \le B$}
        \STATE $E \leftarrow E_1 \cup \textsc{Eval}(h, \mathcal{S}, \{s_2, s_3\})$; \; $\hat{c}_3 \leftarrow \textsc{MeanOutTok}(E)$
        \IF{$\hat{c}_3 \le \gamma_{\mathrm{trig}} B$} \STATE {\bfseries break} \ENDIF
    \ELSE
        \STATE $E \leftarrow E_1$ \hfill $\triangleright$ Fast screen saves $2/3$ search cost
    \ENDIF
    \IF{$k = K$} \STATE {\bfseries break} \ENDIF
    \STATE $h' \leftarrow P\bigl(h,\, \textsc{CostTelemetry}(E),\, \text{target } \gamma_{\mathrm{targ}} B\bigr)$
    \IF{$h'$ invalid} \STATE {\bfseries break} \ENDIF; \; $h \leftarrow h'$
\ENDFOR
\IF{$|E| < 3|\mathcal{S}|$} \STATE $E \leftarrow E \cup \textsc{Eval}(h, \mathcal{S}, \{s_2, s_3\})$ \ENDIF
\STATE Evaluate $h$ on held-out $\mathcal{T}$; record $\hat{C}_{\mathcal{S}}(h)$ and $C(h)$
\end{algorithmic}
\end{algorithm}

\subsection{Measure-Then-Revise Calibration Gate}
\label{sec:method_trim}
Generating a compute router that satisfies a token budget is challenging because its actual cost depends on how frequently its routing rules are triggered during execution. Static execution traces and the costs of previously evaluated candidates provide limited information about the runtime cost of a newly generated router.

The cost difference between reasoning levels is also substantial. In our experiments, high-effort turns consume approximately 673 output tokens, compared with 255 for medium-effort turns, a difference of approximately $2.6\times$.

Our discard-and-resample baseline rejects over-budget drafts and allows up to two replacement attempts, without providing detailed feedback about which routing decisions contributed to the excessive cost. Moreover, routers that operate close to the budget on search tasks may exceed it on unseen tasks.

We introduce \emph{measure-then-revise} (\cref{alg:trim}), which allows the proposer to revise a router using its measured runtime cost rather than discard it. All calibration decisions use only the search split $\mathcal{S}$. The token budget $B$ is defined as twice the root harness's mean output token consumption on search tasks.

The method consists of three steps:
\begin{enumerate}[leftmargin=14pt,itemsep=2.5pt,parsep=0pt,topsep=3pt]
    \item \textbf{Initial cost screening.} We evaluate the router on one rollout of each of the 50 search tasks. If its mean output token consumption exceeds $B$, we trigger revision without completing the remaining two rollouts. This avoids two-thirds of the search evaluation episodes for that draft.
    \item \textbf{Full search evaluation.} If the router passes the initial screen, we complete the remaining rollouts, producing 150 search episodes in total. Revision is triggered if mean token consumption exceeds $0.95B$. This threshold provides a $5\%$ safety margin to reduce the risk of exceeding the budget on unseen tasks.
    \item \textbf{Measurement-guided revision.} When either threshold is exceeded, the proposer receives the router code, measured output token consumption at each reasoning level, the number of turns assigned to each level, and the frequency of regenerations. It then revises the router toward a target of $0.85B$ while attempting to preserve its original reasoning strategy. We allow a maximum of three revisions.
\end{enumerate}

After the revision process ends, the final router completes the search evaluation if necessary and is evaluated on held-out tasks. Held-out observations are not used to revise the router.

Unlike discard-and-resample, measure-then-revise provides the proposer with information about the actual computational behavior of its router, allowing it to make targeted modifications to reduce token consumption. Appendix~\ref{app:prompts} provides the revision prompt and an example of a router before and after calibration.

\section{Experiments}
\label{sec:experiments}

\subsection{Textual Guidelines and Guards Fail to Generalize}
\label{sec:prose}

Across the six experimental settings involving prompt guidelines and pre-write guards, the selected harnesses achieve little or no improvement on held-out tasks, with gains ranging from $-3.0$ to $0.2$ percentage points (pp). On average, candidates do not improve upon their parent harnesses, and the best observed held-out gains are comparable to those obtained without modifying the harness (Appendix~\ref{app:extended}, \cref{tab:prose}).

An analysis of 1,380 root episodes identifies three factors behind these results. First, the original $\tau$-bench retail training tasks always include order IDs and contain no conditional requests, whereas $9\%$ of test tasks omit order IDs and $40\%$ include conditional requests. Second, even on the matched $50/65$ split, rules designed to correct failures improve previously unsuccessful tasks by only $2.3$--$3.3$\,pp while reducing accuracy on previously successful tasks by $26$\,pp. Third, pre-write guards intervene only when the agent attempts a database operation, often after forming and communicating an incorrect plan, and degrade guarded-task performance in 32 of 36 candidates, limiting generalization.

\begin{table*}[t]
\caption{Main results on held-out $\tau$-bench retail tasks.}
\label{tab:main}
\vspace{2pt}
\centering
\resizebox{\textwidth}{!}{%
\begin{tabular}{@{}llrrrrrr@{}}
\toprule
Method / Proposer & Budget Handling & Gain $\Delta_{\mathrm{ship}}$ (pp) & $t_{\mathrm{task}}$ & Cost & $\rho_{\mathrm{start}}$ & Over Cap & Elig.\ Oracle $\Delta_{\mathrm{elig}}^\star$ (pp) \\
\midrule
\multicolumn{8}{@{}l}{\textit{A. Within-Budget Reference (No Loop)}} \\
Uniform \texttt{medium} & Fixed effort tier & \pmsd{9.1}{2.1} & $3.23$ & $1.64\times$ & -- & -- & -- \\
\midrule
\multicolumn{8}{@{}l}{\textit{B. Open-Loop Router Search}} \\
GPT-OSS-120B (self) & No cap control & \pmsd{3.4}{6.4} & $1.86$ & $1.65\times$ & $28/36$ & $27/36$ & \pmsd{5.6}{4.5} \\
GPT-OSS-120B (self) & History + discard & \pmsd{4.8}{1.3} & $2.05$ & $1.88\times$ & $20/36$ & $8/36$ & \pmsd{5.3}{2.1} \\
Gemini 4 Argon & History + discard & \pmsd{7.7}{6.0} & $3.33$ & $1.89\times$ & $34/36$ & $21/36$ & \pmsd{11.1}{4.3} \\
Gemini 3.8 Flash & History + discard & \pmsd{9.1}{5.6} & $3.88$ & $1.93\times$ & $36/36$ & $26/36$ & \pmsd{12.5}{1.2} \\
DeepSeek V4.1 Flash (Vertex) & History + discard & \pmsd{7.9}{2.4} & $2.83$ & $1.79\times$ & $36/36$ & $21/36$ & \pmsd{10.9}{4.6} \\
GLM-5.2 & History + discard & \pmsd{9.2}{6.2} & $3.39$ & $1.86\times$ & $34/36$ & $19/36$ & \pmsd{12.3}{3.6} \\
GLM-5.3 (search cap$^\star$) & History + discard & \pmsd{5.4}{6.4} & $3.10$ & $1.89\times$ & $63/72$ & $38/72$ & \pmsd{9.7}{3.5} \\
GLM-5.3 (combined cap) & History + discard & \pmsd{6.3}{7.1} & $3.69$ & $1.86\times$ & $63/72$ & $43/72$ & \pmsd{9.0}{4.1} \\
\midrule
\multicolumn{8}{@{}l}{\textit{C. Measure-Then-Revise (Ours)}} \\
GLM-5.3 (search cap$^\star$) & History + revise & \pmsd{11.6}{2.9} & $5.74$ & $1.79\times$ & $68/72$ & $2/72$ & \pmsd{11.7}{2.8} \\
GLM-5.3 (combined cap) & History + revise & \pmsd{11.6}{2.9} & $5.74$ & $1.79\times$ & $68/72$ & $4/72$ & \pmsd{11.7}{2.8} \\
DeepSeek V4.1 Flash$^\dagger$ (3 seeds) & History + revise & \pmsd{7.4}{3.6} & $2.70$ & $1.59\times$ & $28/36$ & $0/36$ & \pmsd{11.8}{5.4} \\
\midrule
\multicolumn{8}{@{}l}{\textit{D. Start-From-Medium Negative Controls$^\ddagger$}} \\
Equal absolute cap ($1.2\times$ med.) & History + discard & \pmsd{-1.2}{2.1} & $-1.04$ & $1.01\times$ & $2/36$ & $12/36$ & \pmsd{0.0}{0.0} \\
Loose cap ($2.0\times$ med.) & History + discard & \pmsd{-5.0}{7.8} & $-2.67$ & $1.43\times$ & $6/36$ & $0/36$ & \pmsd{0.9}{1.5} \\
\bottomrule
\end{tabular}%
}\\[2pt]
{\raggedright\scriptsize Mean $\pm$ SD across seeds ($3$ seeds by default, $6$ for GLM-5.3); $t_{\mathrm{task}}$ is task-paired over $n = 65$ held-out retail tasks; Cost is all-episode output tokens vs.\ root. $^\star$Search-only eligibility ($\hat{C}_{\mathcal{S}} \le 2\hat{C}_{\mathcal{S}}(h_0)$); 3-seed rows and ``combined cap'' rows use retrospective $C(h) \le 2C(h_0)$. $^\dagger$Tinker API configuration (\cref{tab:proposers}). $^\ddagger$Candidate-only $\Delta_{\mathrm{cand}}^\star$: $-4.3 \pm 1.6$\,pp ($1.2\times$) and $-1.0 \pm 3.9$\,pp ($2.0\times$).\par}
\vspace{-4pt}
\end{table*}

\subsection{Uniform Reasoning Effort and Event Routers}
\label{sec:headroom}

As a baseline, we evaluate how fixed reasoning effort levels affect \texttt{gpt-oss-120b}. Across all 115 retail test tasks, moving from low to medium effort improves accuracy by \pmsd{6.7}{2.7}\,pp over the $35.9\%$ low-effort baseline at $1.68\times$ output tokens, while uniform high effort gains \pmsd{9.6}{1.9}\,pp at $4.4\times$ tokens. On the 65 held-out tasks, uniform medium achieves \pmsd{9.1}{2.1}\,pp at $1.64\times$ tokens and serves as our within-budget reference (Panel A, \cref{tab:main}). Five event routers targeting write operations, errors, conditional requests, catalog reads, or user turns gain at most \pmsd{1.0}{1.8}\,pp, as escalating isolated triggers misses multi-step order lookups and unnecessary transfers to a human agent (Appendix~\ref{app:extended}).

\subsection{Open-Loop Compute Routers Improve Accuracy but Exceed the Budget}
\label{sec:routers}

\begin{figure*}[t]
\centering
\includegraphics[width=\textwidth]{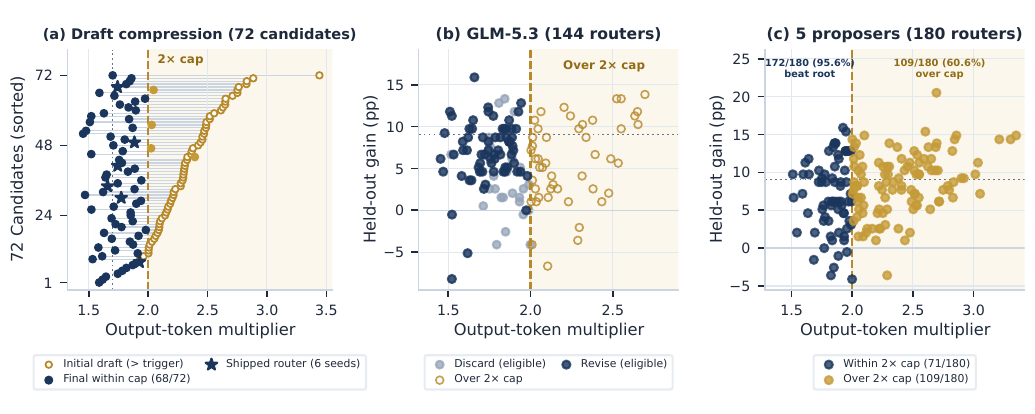}
\vspace{-10pt}
\caption{Candidate router distributions across calibration modes and proposers. (a) Draft-to-revision token compression across all 72 GLM-5.3 revision candidates. (b) Held-out gain vs.\ token multiplier across all 144 GLM-5.3 routers. (c) Balanced 3-seed open-loop slice across five proposers (180 routers; multipliers above $3.35\times$ clipped).}
\label{fig:trim_mech}
\vspace{-4pt}
\end{figure*}

Replacing prompt and guard modifications with executable compute routing produces improvements on held-out tasks, even when the executor itself serves as the proposer (Panel B, \cref{tab:main}). Across five proposers (Gemini 4 Argon, Gemini 3.8 Flash, DeepSeek V4.1 Flash, GLM-5.2, and GLM-5.3), two patterns emerge (\cref{fig:trim_mech}c). First, $95.6\%$ ($172/180$) of generated routers outperform their root harness on held-out tasks, with unconstrained ceilings of $12.0$ to $16.1$\,pp. Second, $60.6\%$ ($109/180$) exceed the $2\times$ token budget despite discarding 49--60 over-budget search drafts per proposer, with within-budget survival falling from $50\%$ in Round 1 to $25\%$ in Round 4 (\cref{fig:generations}b).

The GLM-5.3 comparison further illustrates this bottleneck: although 63 of 72 candidates beat their root harness, 38 exceed the search token cap under the search-only rule ($47.2\%$ eligible; $43/72$ over cap, or $40.3\%$ eligible, under retrospective combined-cost accounting). Under search-only selection, shipped routers average \pmsd{5.4}{6.4}\,pp, with three seeds between $-2.6$ and $1.5$\,pp (\pmsd{6.3}{7.1}\,pp under combined-cost accounting). Removing candidate cost history or discard filtering on GLM-5.2 further degrades selected gains without reducing unconstrained ceilings (Appendix~\ref{app:selection}, \cref{tab:ablation}), showing that retaining effective routers within budget is as critical as generating them.

\begin{figure}[t]
\centering
\includegraphics[width=\columnwidth]{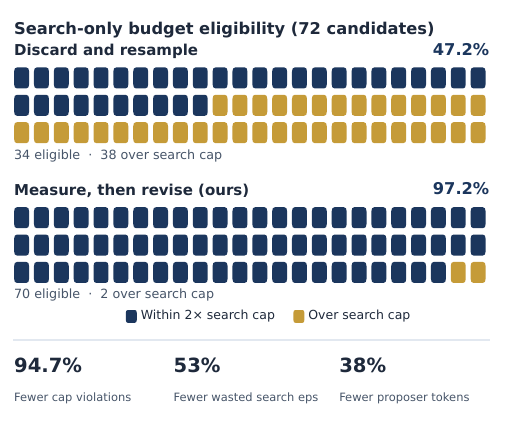}
\vspace{-10pt}
\caption{Search-only budget eligibility and search overhead on GLM-5.3 (72 candidates per method). Each square represents one evaluated candidate router under discard-and-resample (top; $38/72$ over search cap, $43/72$ combined) versus measure-then-revise (bottom; $2/72$ over search cap, $4/72$ combined).}
\label{fig:budget_grid}
\vspace{-6pt}
\end{figure}

\subsection{Closed-Loop Calibration With Measure-Then-Revise}
\label{sec:trim}

\begin{table}[t]
\caption{Per-seed results of \emph{measure-then-revise} on GLM-5.3.}
\label{tab:trim_seeds}
\vspace{2pt}
\centering
\resizebox{\columnwidth}{!}{%
\begin{tabular}{@{}lrrrrrrr@{}}
\toprule
Metric & S0 & S1 & S2 & S3 & S4 & S5 & Mean / Tot \\
\midrule
Root acc.\ (\%) & $36.9$ & $39.0$ & $39.0$ & $39.0$ & $37.4$ & $36.9$ & $38.0$ \\
Shipped gain (pp) & $11.8$ & $6.7$ & $12.3$ & $11.3$ & $15.9$ & $11.8$ & \pmsd{11.6}{2.9} \\
Shipped cost & $1.93\times$ & $1.89\times$ & $1.74\times$ & $1.77\times$ & $1.66\times$ & $1.74\times$ & $1.79 \pm 0.10\times$ \\
Beat root ($\rho_{\mathrm{start}}$) & $12/12$ & $11/12$ & $10/12$ & $11/12$ & $12/12$ & $12/12$ & $68/72$ ($94.4\%$) \\
Over search cap ($\mathcal{E}_{\mathcal{S}}$) & $0/12$ & $1/12$ & $0/12$ & $1/12$ & $0/12$ & $0/12$ & $2/72$ ($2.8\%$) \\
Over combined cap ($\mathcal{E}$) & $0/12$ & $1/12$ & $0/12$ & $2/12$ & $0/12$ & $1/12$ & $4/72$ ($5.6\%$) \\
Over-cap drafts & $16$ & $19$ & $17$ & $10$ & $13$ & $21$ & $96$ total \\
Elig.\ oracle (pp) & $11.8$ & $7.2$ & $12.3$ & $11.3$ & $15.9$ & $11.8$ & \pmsd{11.7}{2.8} \\
\bottomrule
\end{tabular}%
}
\vspace{-6pt}
\end{table}

\subsubsection{Budget Compliance and Search Efficiency}
Measure-then-revise substantially improves budget compliance in the GLM-5.3 experiments. Under the search-only rule, the number of candidates exceeding the $2\times$ token budget decreases from $38/72$ ($52.8\%$) under discard-and-resample to $2/72$ ($2.8\%$) under measure-then-revise (\cref{fig:budget_grid}, \cref{tab:trim_seeds}). Under the retrospective combined-cost rule across all 345 episodes, violations decrease from $43/72$ ($59.7\%$) to $4/72$ ($5.6\%$).

Although 64 of 72 initial drafts trigger revision, with token consumption reaching up to $3.44\times$ that of the root harness, search-only eligibility remains between $94\%$ and $100\%$ across all four search rounds ($89\%$ to $100\%$ under combined-cost accounting).

The method also reduces search overhead. Screening expensive drafts after the first rollout reduces search episodes spent on over-budget drafts by $53\%$, from $13{,}500$ to $6{,}300$ (Appendix~\ref{app:ledger}).

Measure-then-revise selects the same 24 parent harnesses and all six final routers under both $\mathcal{E}_{\mathcal{S}}$ and $\mathcal{E}$, achieving \pmsd{11.6}{2.9}\,pp versus \pmsd{5.4}{6.4}\,pp under discard-and-resample with search-only selection (\pmsd{6.3}{7.1}\,pp under combined-cost accounting; \cref{tab:main}).

These results demonstrate that measure-then-revise improves budget compliance and reduces search overhead without relying on held-out information for calibration or selection.

\subsubsection{Held-Out Accuracy Across Seeds}
All GLM-5.3 measure-then-revise runs produce positive held-out improvements, ranging from $6.7$ to $15.9$\,pp. The selected routers achieve a mean gain of \pmsd{11.6}{2.9}\,pp while consuming $1.79\times$ the baseline output tokens (\cref{tab:trim_seeds}).

In five of the six runs, the selected router matches the best observed held-out performance among all budget-eligible candidates. The mean selection loss is approximately $0.1$\,pp.

Compared with discard-and-resample under the search-only rule, measure-then-revise also produces more consistent outcomes across seeds. The standard deviation of held-out gains decreases from $6.4$ to $2.9$\,pp ($7.1$ to $2.9$\,pp under combined-cost accounting), while the mean number of search-budget violations per seed decreases from $6.3$ to $0.3$ out of 12 candidates ($7.2$ to $0.7$ across all episodes).

The mean improvement over discard-and-resample under the search-only rule is $6.2$\,pp; the seed-level $95\%$ confidence interval ranges from $-0.6$ to $13.0$\,pp, while the task-paired comparison yields $t_{\mathrm{task}} = 3.03$ ($\text{SE} = 2.1$\,pp). Under the retrospective combined-cost rule, the mean difference is $5.3$\,pp (seed-level interval $-2.2$ to $12.8$\,pp; $t_{\mathrm{task}} = 2.73$, $\text{SE} = 1.9$\,pp).

Because baseline accuracy varies across seeds, we also compare absolute held-out accuracy. Under the search-only rule, measure-then-revise achieves \pmsd{49.7}{2.6}\%, compared with \pmsd{44.2}{4.1}\% under discard-and-resample (\pmsd{45.1}{4.4}\% under the retrospective combined-cost rule).

The results favor measure-then-revise in both mean performance and consistency.

\subsubsection{Cross-Proposer Replication and Gate Sensitivity}
To examine whether measure-then-revise extends beyond GLM-5.3, we conduct three additional runs using DeepSeek V4.1 Flash.

Across these runs, none of the 36 evaluated candidates exceeds the token budget. All three selected routers improve upon their root harnesses, achieving a mean held-out gain of \pmsd{7.4}{3.6}\,pp at $1.59\times$ the baseline output tokens.

Because these experiments use a different API provider and decoding configuration than the open-loop DeepSeek experiments (\cref{tab:proposers}), we treat them as a standalone replication rather than a controlled comparison between budget handling methods.

Sensitivity analyses in Appendix~\ref{app:sensitivity} (\cref{tab:gate_sensitivity}) show that allowing up to $K = 3$ revisions lowers GLM-5.3 combined-cost violations from $86.1\%$ ($K = 0$) to $5.6\%$ ($0\%$ on DeepSeek), the $0.95B$ full-search trigger cuts post-search budget drift from $11.9\%$ to $3.3\%$, and the single-rollout screen achieves $92.7\%$ precision ($38/41$).

\begin{figure}[t]
\centering
\includegraphics[width=\columnwidth]{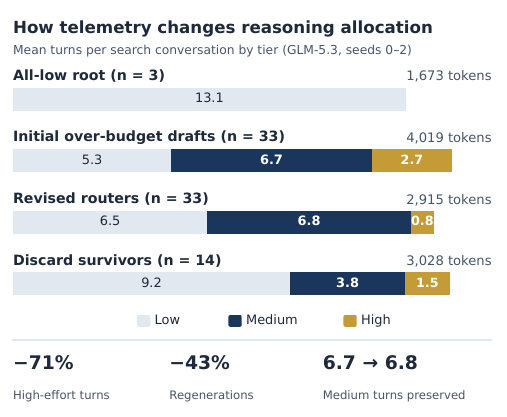}
\vspace{-10pt}
\caption{Per-conversation search effort allocation (GLM-5.3, seeds 0--2). Mean turns per tier and total output tokens across root, initial drafts, revised routers, and discard survivors (\cref{tab:telemetry}).}
\label{fig:allocation}
\vspace{-6pt}
\end{figure}

\subsubsection{Why Measure-Then-Revise Works}
The best observed unconstrained gains are similar under measure-then-revise and discard-and-resample ($12.3$ versus $11.3$\,pp), confirming that the $6.2$\,pp shipped improvement comes from retaining effective routing strategies within budget. Across the first three GLM-5.3 seeds (\cref{fig:allocation}), revision cuts mean high-effort turns by $71\%$ ($2.7$ to $0.8$ per conversation) and regenerations by $43\%$, while preserving medium-effort turns ($6.7$ to $6.8$ versus $3.8$ among discard-and-resample survivors). Across 585 held-out conversations per group, the shipped revised routers reduce unnecessary human hand-offs from 135 to 63 and omitted write operations from 133 to 74, whereas simpler single-rule controls underperform uniform medium effort (Appendix~\ref{app:extended}, \cref{fig:failures,tab:failures}).

\section{Comparison with Uniform-Effort and Learned Compute Allocators}
\label{sec:controls}

\begin{figure}[t]
\centering
\includegraphics[width=\columnwidth]{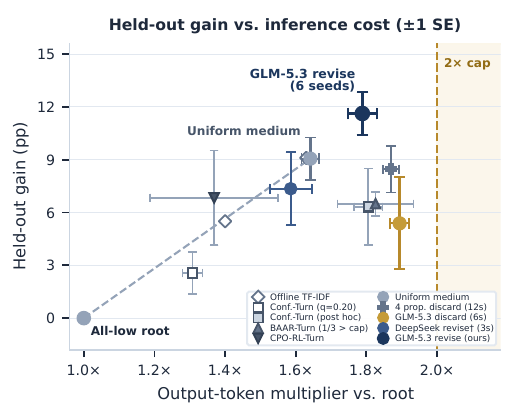}
\vspace{-10pt}
\caption{Held-out gain vs.\ inference cost across loop methods, turn-level learned allocators, and references. Points and error bars show seed means $\pm 1\text{ SE}$ relative to each setting's all-\texttt{low} root; the dashed line connects root and uniform \texttt{medium}.}
\label{fig:frontier}
\vspace{-6pt}
\end{figure}

We compare held-out performance against output token consumption to determine whether harness evolution improves computational efficiency or primarily provides a mechanism for automated budget calibration (\cref{fig:frontier}).

Individual routers discovered through measure-then-revise achieve improvements of up to $15.9$\,pp at $1.66\times$ the baseline token consumption. On GLM-5.3, the method achieves a mean improvement of \pmsd{11.6}{2.9}\,pp at $1.79\times$ tokens. Uniform medium effort achieves \pmsd{9.1}{2.1}\,pp at $1.64\times$ tokens.

A paired comparison across the 65 held-out tasks does not establish a statistically significant difference in absolute accuracy between the two approaches ($49.7\%$ versus $48.2\%$, $t_{\mathrm{task}} = 0.80$). Across search rounds, stronger proposers achieve most of their accuracy improvements in the first round, while subsequent rounds primarily improve budget compliance (Appendix~\ref{app:sensitivity}, \cref{fig:generations}).

\begin{table}[t]
\caption{Turn-level learned allocator re-implementations on held-out retail tasks (3 seeds each).}
\label{tab:learned_allocators}
\vspace{2pt}
\centering
\resizebox{\columnwidth}{!}{%
\begin{tabular}{@{}lrrrrrr@{}}
\toprule
Allocator (Re-Impl.) & S0 & S1 & S2 & Gain (pp) & Cost & $>2\times$ \\
\midrule
Confidence-Turn ($q{=}0.20$) & $2.6$ & $0.5$ & $4.6$ & \pmsd{2.6}{2.1} & $1.31\times$ & $0/3$ \\
Confidence-Turn ($q{=}0.425$, post hoc) & $9.2$ & $2.1$ & $7.7$ & \pmsd{6.3}{3.8} & $1.81\times$ & $0/3$ \\
BAAR-Turn (ModernBERT) & $5.1$ & $7.2^\dagger$ & $7.2$ & \pmsd{6.5}{1.2} & $1.83\times$ & $1/3$ \\
CPO-RL-Turn (primal-dual) & $11.8$ & $6.2$ & $2.6$ & \pmsd{6.8}{4.7} & $1.37\times$ & $0/3$ \\
\bottomrule
\end{tabular}%
}\\[2pt]
{\raggedright\scriptsize Gains in pp over each seed's all-\texttt{low} root ($65$ held-out tasks $\times 3$ rollouts). $q{=}0.425$ bisected on search after viewing $q{=}0.20$ held-out results. $^\dagger$Seed 1 spent $1.74\times$ on search and $2.04\times$ on held-out.\par}
\vspace{-6pt}
\end{table}

We also compare against three reimplemented turn-level compute allocators calibrated on the same 50 search tasks (\cref{tab:learned_allocators}; Appendix~\ref{app:extended}). Confidence-Turn uses token log probabilities, gaining \pmsd{2.6}{2.1}\,pp at $1.31\times$ tokens on its predefined grid ($q = 0.20$) and \pmsd{6.3}{3.8}\,pp at $1.81\times$ under post hoc exploratory calibration ($q = 0.425$). BAAR-Turn \citep{jali2026,zhang2026baar} trains ModernBERT on $2{,}400$ search episodes ($31{,}948$ turns) and gains \pmsd{6.5}{1.2}\,pp at $1.83\times$ tokens, with modest validation AUC ($0.58$--$0.60$) and one seed exceeding cap on held-out tasks ($2.04\times$). CPO-RL-Turn \citep{yang2026ares,zhai2026} trains a primal-dual policy on $2{,}400$ search episodes per seed and gains \pmsd{6.8}{4.7}\,pp at $1.37\times$ tokens, though its three seeds collapse toward uniform medium, the initial policy, and uniform low rather than consistent state-dependent routing.

Two additional controls highlight the limits of automated routing: starting evolution from uniform medium effort degrades accuracy under both tested caps (Panel D, \cref{tab:main}), and an offline TF-IDF opening-message classifier tracks the same cost--accuracy trend as uniform medium (\cref{fig:frontier}; \cref{tab:offline_router}). Thus, the main benefit of harness evolution in our setting is autonomously discovering and calibrating effective effort allocation without weight training, rather than surpassing uniform medium efficiency.

Additional analyses of candidate selection, domain transfer, and executor scope are reported in Appendices~\ref{app:selection} and~\ref{app:partitions}. Re-scoring eligible routers with $12$ rather than three search rollouts recovers $22\%$ of the remaining selection gap (\cref{fig:selection,tab:rescore_full}), while open-loop router search on $\tau$-bench airline reproduces both the held-out accuracy gains ($34/36$ beat root; \pmsd{5.2}{5.1}\,pp shipped) and budget violations ($27/36$ over cap) observed in retail. Uniform-effort probes on three additional executors (\texttt{Qwen3.8-27B}, \texttt{Inkling 2 Small}, and \texttt{gpt-oss-20b}; \cref{tab:exec2}) show no statistically significant improvements from higher uniform effort, suggesting that adaptive effort routing requires an executor whose accuracy responds to increased reasoning compute.

\section{Conclusion}
\label{sec:conclusion}

Evaluating harness self-improvement on unseen tasks under a fixed token budget shows that both the choice of harness modifications and runtime cost control are decisive. Unlike prompt and guard edits, executable compute routers generalize across proposer models, yet open-loop generation routinely exceeds the budget. \emph{Measure-then-revise} addresses this bottleneck by feeding per-tier token measurements back into revision, eliminating nearly all budget violations, halving wasted search evaluations, and yielding consistent held-out gains across seeds. Because evolved routers match rather than surpass uniform medium effort, harness evolution here acts primarily as automated, training-free compute calibration, motivating extensions to multi-model routing and stronger test-time-scaling executors.

{\small\linespread{0.96}\selectfont
\setlength{\bibsep}{1.2pt plus 0.2pt}
\bibliography{refs}

@inproceedings{agrawal2025,
  author = {Lakshya A Agrawal and Shangyin Tan and Dilara Soylu and Noah Ziems and Rishi Khare and Krista Opsahl-Ong and Arnav Singhvi and Herumb Shandilya and Michael J Ryan and Meng Jiang and others},
  title = {{GEPA: Reflective Prompt Evolution Can Outperform Reinforcement Learning}},
  booktitle = {International Conference on Learning Representations (ICLR)},
  year = {2026},
  url = {https://arxiv.org/abs/2507.19457}
}

@article{brown2024,
  author = {Bradley Brown and Jordan Juravsky and Ryan Ehrlich and Ronald Clark and Quoc V. Le and Christopher R\'{e} and Azalia Mirhoseini},
  title = {{Large Language Monkeys: Scaling Inference Compute with Repeated Sampling}},
  journal = {arXiv preprint arXiv:2407.21787},
  year = {2024},
  url = {https://arxiv.org/abs/2407.21787}
}

@article{chen2026,
  author = {Chen, Zihan and Zhu, Di and Zheng, Lei Nico},
  title = {{Auditing the Product, Not the Recursion: The Evaluation Evidence Behind Recursive Self-Improvement Claims}},
  journal = {arXiv preprint arXiv:2610.27610},
  year = {2026},
  url = {https://arxiv.org/abs/2610.27610}
}

@article{he2026,
  author = {Rong He},
  title = {{Does Failure Feedback Direct Agent Self-Modification?}},
  journal = {arXiv preprint arXiv:2610.14662},
  year = {2026},
  url = {https://arxiv.org/abs/2610.14662}
}

@article{ding2026,
  author = {Wenxuan Ding and Qin Lu and Changlong Yu and Sha Li and Shuowei Jin and Xin Liu and Zhenyu Shi and Greg Durrett},
  title = {{Harness-Delta Attribution: Dissecting Gains From Harness Evolution}},
  journal = {arXiv preprint arXiv:2610.22466},
  year = {2026},
  url = {https://arxiv.org/abs/2610.22466}
}

@article{tian2026,
  author = {Tian, Yuan and Hu, Bing and Wang, Hao and Lu, Binghang and Wu, Fang},
  title = {{Harness Evolution Hits a Ceiling: When Weight Training Should Begin}},
  journal = {arXiv preprint arXiv:2610.11655},
  year = {2026},
  url = {https://arxiv.org/abs/2610.11655}
}

@article{wei2026,
  author = {Chuyang Wei and Haoxiang Guan and Minghang Zhu and Kexin Zhou and Jiyan He and Xiren Zhou and Huanhuan Chen and Yifei Shen},
  title = {{Self-Evolving Agent Harness}},
  journal = {arXiv preprint arXiv:2610.18813},
  year = {2026},
  url = {https://arxiv.org/abs/2610.18813}
}

@article{luo2026,
  author = {Haochen Luo and Yi Huang and Sichun Luo and Fengyuan Liu and Lei Li and Zefa Hu and Junlan Feng and Qi Liu},
  title = {{Harness-Aware Self-Evolving: Co-Evolving Model Weights, Harness, and Task Solutions}},
  journal = {arXiv preprint arXiv:2607.03935},
  year = {2026},
  url = {https://arxiv.org/abs/2607.03935}
}

@article{sengupta2026,
  author = {Biswa Sengupta and Jinhua Wang},
  title = {{HARBOR: Automated Harness Optimization}},
  journal = {arXiv preprint arXiv:2604.20938},
  year = {2026},
  url = {https://arxiv.org/abs/2604.20938}
}

@article{he2026b,
  author = {Liang He and Jingbo Wen and Haoyu Wang and Ziqi He and Yixiong Chen and Kangning Cui and Xilu Wang},
  title = {{Not All Errors Are Equal: Consequence-Aware Reasoning Compute Allocation}},
  journal = {arXiv preprint arXiv:2606.04402},
  year = {2026},
  url = {https://arxiv.org/abs/2606.04402}
}

@inproceedings{hu2024,
  author = {Shengran Hu and Cong Lu and Jeff Clune},
  title = {{Automated Design of Agentic Systems}},
  booktitle = {International Conference on Learning Representations (ICLR)},
  year = {2025},
  url = {https://arxiv.org/abs/2408.08435}
}

@article{jali2026,
  author = {Neharika Jali and Anupam Nayak and Gauri Joshi},
  title = {{Not All Turns Are Equally Hard: Adaptive Thinking Budgets For Efficient Multi-Turn Reasoning in Agents}},
  journal = {arXiv preprint arXiv:2604.05164},
  year = {2026},
  url = {https://arxiv.org/abs/2604.05164}
}

@article{kassenaar2026,
  author = {Gijs Kassenaar and Zhao Yang and Vincent Fran{\c{c}}ois-Lavet},
  title = {{Learning When to Think: Adaptive Reasoning for Test-Time Compute Allocation}},
  journal = {arXiv preprint arXiv:2608.20256},
  year = {2026},
  url = {https://arxiv.org/abs/2608.20256}
}

@article{lee2026,
  author = {Woongkyu Lee and Jungwook Choi},
  title = {{Rethinking Inference-Time Scaling in Local Computer-Use Agents: Failure Modes and Compute Tradeoffs}},
  journal = {arXiv preprint arXiv:2607.28573},
  year = {2026},
  url = {https://arxiv.org/abs/2607.28573}
}

@article{madiraju2026,
  author = {Meher Bhaskar Madiraju and Meher Sai Preetam Madiraju},
  title = {{AgentSLABench: Evaluating and Benchmarking Agentic Systems Under Resource Constraints}},
  journal = {arXiv preprint arXiv:2608.00805},
  year = {2026},
  url = {https://arxiv.org/abs/2608.00805}
}

@article{sah2026,
  author = {Chandan Kumar Sah and Li Zhang and Xiaoli Lian},
  title = {{Beyond Self-Knowledge: Propagating Uncertainty Across Reasoning and Retrieval in LLMs}},
  journal = {arXiv preprint arXiv:2607.25600},
  year = {2026},
  url = {https://arxiv.org/abs/2607.25600}
}

@article{zhang2026,
  author = {Jenny Zhang and Bingchen Zhao and Wannan Yang and Jakob Foerster and Jeff Clune and Minqi Jiang and Sam Devlin and Tatiana Shavrina},
  title = {{Hyperagents}},
  journal = {arXiv preprint arXiv:2603.19461},
  year = {2026},
  url = {https://arxiv.org/abs/2603.19461}
}

@article{kapoor2024,
  author = {Kapoor, Sayash and Stroebl, Benedikt and Siegel, Zachary S. and Nadgir, Nitya and Narayanan, Arvind},
  title = {{AI Agents That Matter}},
  journal = {Transactions on Machine Learning Research (TMLR)},
  year = {2024},
  url = {https://arxiv.org/abs/2407.01502}
}

@article{lee2026b,
  author = {Yoonho Lee and Roshen Nair and Qizheng Zhang and Kangwook Lee and Omar Khattab and Chelsea Finn},
  title = {{Meta-Harness: End-to-End Optimization of Model Harnesses}},
  journal = {arXiv preprint arXiv:2603.28052},
  year = {2026},
  url = {https://arxiv.org/abs/2603.28052}
}

@article{shang2026,
  author = {Zhiwei Shang and Yu Huo and Mingrong Gong and An Yan and Zikun Qu and Junhao Dong and Bryan Kian Hsiang Low and Chenglin Wu and Zhongxiang Dai},
  title = {{MESH-Harness: Self-Improving Agent Harnesses via Bandit-Guided Compositional Evolution}},
  journal = {arXiv preprint arXiv:2610.05300},
  year = {2026},
  url = {https://arxiv.org/abs/2610.05300}
}

@article{miller2024,
  author = {Miller, Evan},
  title = {{Adding Error Bars to Evals: A Statistical Approach to Language Model Evaluations}},
  journal = {arXiv preprint arXiv:2411.00640},
  year = {2024},
  url = {https://arxiv.org/abs/2411.00640}
}

@article{xie2026,
  author = {Xie, Zixuan and Xu, Chang and Jiang, Yuhua and Zhao, Li},
  title = {{Predict Before You Modify: Learning Improvement Dynamics for Self-Evolving Agents}},
  journal = {arXiv preprint arXiv:2610.13991},
  year = {2026},
  url = {https://arxiv.org/abs/2610.13991}
}

@article{ding2026b,
  author = {Tianyu Ding and Yaocheng Tong and Muzhi Yang and Fei Jiang and Lei Yang and Boya Yu},
  title = {{Re-executing Coding-Agent Evaluations: Unchanged Decisions and Unresolved Policy Risk}},
  journal = {arXiv preprint arXiv:2610.14460},
  year = {2026},
  url = {https://arxiv.org/abs/2610.14460}
}

@article{wang2026,
  author = {Wang, Yike and Zhu, Huaisheng and Hu, Zhengyu and Yuan, Yige and Chen, Zhengyu and Senthil, Shakti and Hajishirzi, Hannaneh and Tsvetkov, Yulia and Smith, Noah A. and Dasigi, Pradeep and Xiao, Teng},
  title = {{Rethinking the Evaluation of Harness Evolution for Agents}},
  journal = {arXiv preprint arXiv:2607.12227},
  year = {2026},
  url = {https://arxiv.org/abs/2607.12227}
}

@article{robeyns2025,
  author = {Maxime Robeyns and Martin Szummer and Laurence Aitchison},
  title = {{A Self-Improving Coding Agent}},
  journal = {arXiv preprint arXiv:2504.15228},
  year = {2025},
  url = {https://arxiv.org/abs/2504.15228}
}

@inproceedings{snell2024,
  author = {Charlie Snell and Jaehoon Lee and Kelvin Xu and Aviral Kumar},
  title = {{Scaling LLM Test-Time Compute Optimally can be More Effective than Scaling Model Parameters}},
  booktitle = {International Conference on Learning Representations (ICLR)},
  year = {2025},
  url = {https://arxiv.org/abs/2408.03314}
}

@inproceedings{wang2025,
  author = {Wenyi Wang and Piotr Piekos and Li Nanbo and Firas Laakom and Yimeng Chen and Mateusz Ostaszewski and Mingchen Zhuge and J\"{u}rgen Schmidhuber},
  title = {{Huxley-G\"{o}del Machine: Human-Level Coding Agent Development by an Approximation of the Optimal Self-Improving Machine}},
  booktitle = {International Conference on Learning Representations (ICLR)},
  year = {2026},
  url = {https://arxiv.org/abs/2510.21614}
}

@inproceedings{yao2024,
  author = {Yao, Shunyu and Shinn, Noah and Razavi, Pedram and Narasimhan, Karthik},
  title = {{$\tau$-bench: A Benchmark for Tool-Agent-User Interaction in Real-World Domains}},
  booktitle = {International Conference on Learning Representations (ICLR)},
  year = {2025},
  url = {https://arxiv.org/abs/2406.12045}
}

@inproceedings{yin2024,
  author = {Xunjian Yin and Xinyi Wang and Liangming Pan and Li Lin and Xiaojun Wan and William Yang Wang},
  title = {{G\"{o}del Agent: A Self-Referential Agent Framework for Recursive Self-Improvement}},
  booktitle = {Proceedings of the 63rd Annual Meeting of the Association for Computational Linguistics (ACL)},
  year = {2025},
  url = {https://arxiv.org/abs/2410.04444}
}

@inproceedings{zelikman2023,
  author = {Eric Zelikman and Eliana Lorch and Lester Mackey and Adam Tauman Kalai},
  title = {{Self-Taught Optimizer (STOP): Recursively Self-Improving Code Generation}},
  booktitle = {First Conference on Language Modeling (COLM)},
  year = {2024},
  url = {https://arxiv.org/abs/2310.02304}
}

@inproceedings{zhang2024,
  author = {Jiayi Zhang and Jinyu Xiang and Zhaoyang Yu and Fengwei Teng and Xionghui Chen and Jiaqi Chen and Mingchen Zhuge and Xin Cheng and Sirui Hong and Jinlin Wang and others},
  title = {{AFlow: Automating Agentic Workflow Generation}},
  booktitle = {International Conference on Learning Representations (ICLR)},
  year = {2025},
  url = {https://arxiv.org/abs/2410.10762}
}

@inproceedings{zhang2025,
  author = {Jenny Zhang and Shengran Hu and Cong Lu and Robert Lange and Jeff Clune},
  title = {{Darwin G\"{o}del Machine: Open-Ended Evolution of Self-Improving Agents}},
  booktitle = {International Conference on Learning Representations (ICLR)},
  year = {2026},
  url = {https://arxiv.org/abs/2505.22954}
}

@inproceedings{yang2023,
  author = {Chengrun Yang and Xuezhi Wang and Yifeng Lu and Hanxiao Liu and Quoc V. Le and Denny Zhou and Xinyun Chen},
  title = {{Large Language Models as Optimizers}},
  booktitle = {International Conference on Learning Representations (ICLR)},
  year = {2024},
  url = {https://arxiv.org/abs/2309.03409}
}

@inproceedings{guo2023,
  author = {Qingyan Guo and Rui Wang and Junliang Guo and Bei Li and Kaitao Song and Xu Tan and Guoqing Liu and Jiang Bian and Yujiu Yang},
  title = {{EvoPrompt: Connecting LLMs with Evolutionary Algorithms Yields Powerful Prompt Optimizers}},
  booktitle = {International Conference on Learning Representations (ICLR)},
  year = {2024},
  url = {https://arxiv.org/abs/2309.08532}
}

@inproceedings{fernando2023,
  author = {Chrisantha Fernando and Dylan Banarse and Henryk Michalewski and Simon Osindero and Tim Rockt\"{a}schel},
  title = {{Promptbreeder: Self-Referential Self-Improvement Via Prompt Evolution}},
  booktitle = {International Conference on Machine Learning (ICML)},
  year = {2024},
  url = {https://arxiv.org/abs/2309.16797}
}

@inproceedings{khattab2023,
  author = {Omar Khattab and Arnav Singhvi and Paridhi Maheshwari and Zhiyuan Zhang and Keshav Santhanam and Sri Vardhamanan and Saiful Haq and Ashutosh Sharma and Thomas T. Joshi and Hanna Moazam and others},
  title = {{DSPy: Compiling Declarative Language Model Calls into Self-Improving Pipelines}},
  booktitle = {International Conference on Learning Representations (ICLR)},
  year = {2024},
  url = {https://arxiv.org/abs/2310.03714}
}

@inproceedings{opsahlong2024,
  author = {Krista Opsahl-Ong and Michael J Ryan and Josh Purtell and David Broman and Christopher Potts and Matei Zaharia and Omar Khattab},
  title = {{Optimizing Instructions and Demonstrations for Multi-Stage Language Model Programs}},
  booktitle = {Conference on Empirical Methods in Natural Language Processing (EMNLP)},
  year = {2024},
  url = {https://arxiv.org/abs/2406.11695}
}

@article{yuksekgonul2024,
  author = {Mert Yuksekgonul and Federico Bianchi and Joseph Boen and Sheng Liu and Pan Lu and Zhi Huang and Carlos Guestrin and James Zou},
  title = {{Optimizing generative AI by backpropagating language model feedback}},
  journal = {Nature},
  volume = {639},
  pages = {609--616},
  year = {2025},
  url = {https://arxiv.org/abs/2406.07496}
}

@inproceedings{cheng2024,
  author = {Ching-An Cheng and Allen Nie and Adith Swaminathan},
  title = {{Trace is the Next AutoDiff: Generative Optimization with Rich Feedback, Execution Traces, and LLMs}},
  booktitle = {Advances in Neural Information Processing Systems (NeurIPS)},
  year = {2024},
  url = {https://arxiv.org/abs/2406.16218}
}

@article{gema2025,
  author = {Aryo Pradipta Gema and Alexander H\"{a}gele and Runjin Chen and Andy Arditi and Jacob Goldman-Wetzler and Kit Fraser-Taliente and Henry Sleight and Linda Petrini and Julian Michael and Beatrice Alex and Pasquale Minervini and Yanda Chen and Joe Benton and Ethan Perez},
  title = {{Inverse Scaling in Test-Time Compute}},
  journal = {arXiv preprint arXiv:2507.14417},
  year = {2025},
  url = {https://arxiv.org/abs/2507.14417}
}

@inproceedings{song2024,
  author = {Yuda Song and Hanlin Zhang and Carson Eisenach and Sham Kakade and Dean Foster and Udaya Ghai},
  title = {{Mind the Gap: Examining the Self-Improvement Capabilities of Large Language Models}},
  booktitle = {International Conference on Learning Representations (ICLR)},
  year = {2025},
  url = {https://arxiv.org/abs/2412.02674}
}

@inproceedings{huang2024,
  author = {Audrey Huang and Adam Block and Dylan J. Foster and Dhruv Rohatgi and Cyril Zhang and Max Simchowitz and Jordan T. Ash and Akshay Krishnamurthy},
  title = {{Self-Improvement in Language Models: The Sharpening Mechanism}},
  booktitle = {International Conference on Learning Representations (ICLR)},
  year = {2025},
  url = {https://arxiv.org/abs/2412.01951}
}

@article{yang2026ares,
  author = {Jingbo Yang and Bairu Hou and Wei Wei and Yujia Bao and Shiyu Chang},
  title = {{Ares: Adaptive Reasoning Effort Selection for Efficient LLM Agents}},
  journal = {arXiv preprint arXiv:2603.07915},
  year = {2026},
  url = {https://arxiv.org/abs/2603.07915}
}

@article{zhang2026baar,
  author = {Caiqi Zhang and Menglin Xia and Xuchao Zhang and Daniel Madrigal and Ankur Mallick and Samuel Kessler and Victor Ruehle and Saravan Rajmohan},
  title = {{Budget-Aware Agentic Routing via Boundary-Guided Training}},
  journal = {arXiv preprint arXiv:2602.21227},
  year = {2026},
  url = {https://arxiv.org/abs/2602.21227}
}

@article{jiang2026harnessmm,
  author = {Yuhua Jiang and Chang Xu and Feifei Gao and Li Zhao},
  title = {{HarnessMM: Recursive Self-Improvement for Harness Token Efficiency}},
  journal = {arXiv preprint arXiv:2610.32813},
  year = {2026},
  url = {https://arxiv.org/abs/2610.32813}
}

@article{zhang2026compgen,
  author = {Alex L. Zhang and Souradip Chakraborty and Omar Khattab},
  title = {{Language Model Harnesses Can Be Compositional Generalizers}},
  journal = {arXiv preprint arXiv:2611.28298},
  year = {2026},
  url = {https://arxiv.org/abs/2611.28298}
}

@article{zhai2026,
  author = {Zhiyuan Zhai and Bingcong Li and Bingnan Xiao and Ming Li and Xin Wang},
  title = {{Adaptive Test-Time Compute Allocation for Reasoning LLMs via Constrained Policy Optimization}},
  journal = {arXiv preprint arXiv:2604.14853},
  year = {2026},
  url = {https://arxiv.org/abs/2604.14853}
}

@article{hong2026,
  author = {Zhi Hong and Qianlong Wang and Hongchen Wei and Aidan Wu and Daohong Ying and Mingze Kong and Bei Liu and Chong Luo},
  title = {{HarnessFlow: MCTS-Guided Harness Optimization for LLM Agents}},
  journal = {arXiv preprint arXiv:2611.06254},
  year = {2026},
  url = {https://arxiv.org/abs/2611.06254}
}

@article{ling2026,
  author = {Chen Ling and Pei Chen and Albert Guan and Jiaming Qu and Shayan Ali Akbar and Madhu Gopinathan and Erwin Cornejo},
  title = {{PACE: Two-Timescale Self-Evolution for Small Language Model Agents}},
  journal = {arXiv preprint arXiv:2605.23019},
  year = {2026},
  url = {https://arxiv.org/abs/2605.23019}
}

@article{zhang2026cig,
  author = {Junjie Zhang and Shunyu Liu and Haoyu Wang and Ting-En Lin and Yongbin Li and Dacheng Tao},
  title = {{Causal Improvement Graph for Agentic Harness Optimization}},
  journal = {arXiv preprint arXiv:2610.05039},
  year = {2026},
  url = {https://arxiv.org/abs/2610.05039}
}
\bibliographystyle{plainnat}
}

\clearpage
\appendix
\onecolumn
\raggedbottom
\renewcommand{\topfraction}{0.95}
\renewcommand{\bottomfraction}{0.85}
\renewcommand{\textfraction}{0.05}
\renewcommand{\floatpagefraction}{0.80}
\makeatletter\setlength{\@fptop}{0pt}\setlength{\@fpsep}{12pt}\makeatother

\section{Experimental Setup, Infrastructure, and Complete Experiment Ledger}
\label{app:ledger}

\subsection{Serving Infrastructure, Search Configuration, and Validity Checks}
\label{app:serving}

In the main experiments, both the executor and the customer simulator use the open-weight model \texttt{openai/gpt-oss-120b}, served via vLLM on a single NVIDIA H100 80\,GB GPU.

We launch vLLM using the following configuration:

\begin{listingcard}{Listing 1. vLLM serving configuration}
\begin{lstlisting}[style=bashcard]
vllm serve openai/gpt-oss-120b \
  --served-model-name gpt-oss-120b \
  --gpu-memory-utilization 0.90 \
  --max-model-len 32768 \
  --enable-prefix-caching \
  --enable-auto-tool-choice \
  --tool-call-parser openai
\end{lstlisting}
\end{listingcard}

We set \texttt{VLLM\_USE\_DEEP\_GEMM=0} and \texttt{VLLM\_USE\_FLASHINFER\_SAMPLER=0}, with concurrency ranging from 24 to 32.

Our execution wrapper specifies the reasoning effort for each executor call. The executor uses a temperature of $0.0$ and a maximum of $4{,}096$ output tokens per turn. The customer simulator uses a temperature of $1.0$.

In the standard retail configuration, each independent loop starts from a root harness and runs four search rounds, generating three candidates per round. Including the root, a complete loop evaluates 13 harnesses.

Each harness is evaluated on 150 search episodes and 195 held-out episodes, resulting in 345 conversations per harness and $4{,}485$ conversations per complete loop. These counts exclude additional search episodes used to evaluate rejected or replaced drafts.

We perform four automated validity checks on all runs:
\begin{enumerate}[leftmargin=18pt,itemsep=2pt,parsep=0pt,topsep=3pt]
    \item Verify that each evaluated harness contains the expected number of search and held-out episodes.
    \item Confirm that search and held-out task IDs do not overlap.
    \item Check for unhandled Python exceptions.
    \item Verify that evaluated, rejected, and revised routers are archived together with their per-episode token counts.
\end{enumerate}

Across the GLM-5.3 measure-then-revise runs, the numbers of episode errors are $0, 0, 1, 0, 1$, and $0$, respectively. No router exceptions occur. Across all three DeepSeek measure-then-revise runs, we observe neither episode errors nor router exceptions. Transient HTTP 500 harmony-parser responses from \texttt{gpt-oss-120b} are retried automatically; any unhandled episode exception is retained in the fixed evaluation denominator ($150$ search or $195$ held-out episodes) with reward $0.0$ rather than discarded or redrawn.

\subsubsection{Search Overhead, Evaluation Reuse, and Token Accounting}

Across the 72 GLM-5.3 measure-then-revise candidates, 64 initial drafts trigger budget revision. In total, 96 draft evaluations trigger revision: 81 during the initial single-rollout screen and 15 during the full-search evaluation at the $0.95B$ threshold.

Of these revision attempts, 94 produce valid revisions across 63 candidates. In the remaining two cases, the proposer does not return a valid revision, and the previously evaluated router is retained.

Measure-then-revise uses $6{,}300$ search episodes to evaluate drafts that trigger revision, compared with $13{,}500$ under discard-and-resample, a reduction of $53\%$. Of these $6{,}300$ episodes, 100 are reused in subsequent evaluation. Excluding reused episodes increases the reduction to $54\%$.

Across GLM-5.3 seeds, the wall-clock time spent evaluating over-budget drafts decreases from $23.6$ to $13.7$ hours, a reduction of $42\%$.

In the three DeepSeek V4.1 Flash measure-then-revise runs, the method performs 43 revisions, including 34 triggered by the initial screen and nine triggered by the full-search threshold. These evaluations consume $3{,}050$ search episodes. All 36 final candidates satisfy the budget, and 28 outperform their root harnesses.

Although our computational budget is defined using executor output tokens, we also examine prompt tokens and total token consumption. The selected routers generated by stronger proposers consume $1.03$--$1.33\times$ the root harness's prompt tokens and $1.04$--$1.35\times$ its total tokens, with a mean total-token multiplier of $1.21\times$. The relative method rankings remain unchanged under this alternative accounting.

\subsection{Proposer Models and Reproduction Instructions}

\Cref{tab:proposers} reports the exact proposer model identifiers, serving providers, reasoning-effort configurations, maximum output token limits, number of independent seeds, and token accounting notes. All proposer models use a temperature of $1.0$.

\begin{table}[!ht]
\centering
\caption{Summary of proposer endpoints, serving paths, decoding overrides (all temp $1.0$), and token accounting across settings.}
\label{tab:proposers}
\vspace{3pt}
\resizebox{\textwidth}{!}{%
\begin{tabular}{@{}llllrrl@{}}
\toprule
Proposer Label & Exact Model Identifier & Serving Path & Effort & \texttt{max\_tokens} & Seeds & Token / Runtime Accounting \\
\midrule
GPT-OSS-120B (self) & \texttt{openai/gpt-oss-120b} & H100 (vLLM) & \texttt{medium} & $12{,}000$ & $3$ & $\sim\!11$--$15$ loop wall-clock h/seed \\
Gemini 4 Argon & \texttt{google/gemini-4-argon} & Vertex AI & default & $12{,}000$ & $3$ & standard completion accounting \\
Gemini 3.8 Flash & \texttt{google/gemini-3.8-flash} & Vertex AI & default & $32{,}000$ & $3$ & $\sim\!17.9\text{k}$ thinking tok/call \\
DeepSeek V4.1 Flash (Vertex) & \texttt{deepseek-ai/deepseek-v4.1-flash-maas} & Vertex AI & default & $12{,}000$ & $3$ & $36$ $>150$-line drafts resampled \\
GLM-5.2 & \texttt{zai-org/glm-5.2-maas} & Vertex AI & default & $65{,}536$ & $3$ & reasoning included in completion tok \\
GLM-5.3 (discard) & \texttt{zai-org/GLM-5.3:peft:262144} & Tinker & \texttt{high} & $12{,}000$ & $6$ & $2.27\text{M}$ in / $2.95\text{M}$ out (eval+rej) \\
GLM-5.3 (revise) & \texttt{zai-org/GLM-5.3:peft:262144} & Tinker & \texttt{high} & $12{,}000$ & $6$ & $1.34\text{M}$ in / $1.90\text{M}$ out (eval incl.\ rev) \\
DeepSeek V4.1 Flash (revise) & \texttt{deepseek-ai/DeepSeek-V4.1-Flash:peft:262144} & Tinker & \texttt{low} & $40{,}000$ & $3$ & $386{,}637$ in / $777{,}758$ out (eval) \\
\bottomrule
\end{tabular}%
}
\end{table}

\subsubsection{Reproducing the GLM-5.3 Comparison}

This comparison evaluates discard-and-resample and measure-then-revise using six independent loop seeds per method, with the executor, task splits, search configuration, and output-token budget held fixed.

\noindent\textbf{Step 1. Initialize the execution environment.}

Launch the executor and customer simulator using the vLLM configuration described in Appendix~\ref{app:serving} (endpoint \codepill{localhost:8000/v1}, with \codepill{USER\_MODEL=gpt-oss-120b}).

Configure the proposer environment:

\begin{listingcard}{Listing 2. Environment configuration}
\begin{lstlisting}[style=bashcard]
export TINKER_API_KEY=<YOUR_API_KEY>
export TINKER_REASONING=high
export TINKER_MAX_TOKENS=12000
\end{lstlisting}
\end{listingcard}

\noindent\textbf{Step 2. Execute the independent search loops.}

Run both methods across six independent seeds, using the following shared configuration:

\begin{listingcard}{Listing 3. Shared experimental configuration (seeds 0--5)}
\begin{lstlisting}[style=bashcard]
for s in 0 1 2 3 4 5; do
  python3 loop.py \
    --proposer tinker-zai-org/GLM-5.3:peft:262144 \
    --executor gpt-oss-120b \
    --feedback diff --edit-space route --seed "$s" \
    --env retail --split-mode test-partition --partition-seed 4321 \
    --nodes 13 --children-per-gen 3 \
    --search-k 50 --search-rollouts 3 \
    --gold-k 65 --gold-rollouts 3 \
    --token-budget-mult 2.0 --history --concurrency 24 \
    --out "runs/seed_${s}"
done
\end{lstlisting}
\end{listingcard}

\noindent Append the appropriate argument for each budget handling method:

\vspace{4pt}
\noindent
{\small
\renewcommand{\arraystretch}{1.45}
\arrayrulecolor{cardborder}
\begin{tabular}{@{}p{0.48\textwidth}p{0.48\textwidth}@{}}
Method & Additional argument \\
\midrule
Discard-and-resample & \codepill{-{}-budget-repair 2} \\
\midrule
Measure-then-revise & \codepill{-{}-trim 3} \\
\end{tabular}
\arrayrulecolor{black}
}
\vspace{8pt}

\noindent\textbf{Step 3. Aggregate results and verify experimental accounting.}

After completing all runs, execute the following scripts to reproduce the aggregate results, task-paired comparisons, and search-overhead measurements:

\begin{listingcard}{Listing 4. Summary and overhead audit}
\begin{lstlisting}[style=bashcard]
python3 compare_revise_vs_discard.py
python3 audit_6seed.py
\end{lstlisting}
\end{listingcard}

\subsection{Complete Experiment Ledger}

\Cref{tab:full_ledger} summarizes all reported experimental settings, including the corresponding result files, evaluation splits, number of seeds, selected harness gains, task-paired statistics, candidate improvement rates, budget violations, and oracle gains.

Gains and oracle values are reported in decimal reward units. The task-paired statistic, $t_{\mathrm{task}}$, uses 65 paired tasks for retail, 30 for airline, and 115 for the original train-to-test experiments. The pooled statistic, $t_{\mathrm{pool}}$, treats task--seed combinations as paired observations.

Our analysis reports task-paired statistics alongside performance variability across independent loop seeds. GLM-5.3 archives store per-episode split logs enabling search-only eligibility $\mathcal{E}_{\mathcal{S}}$, whereas multi-proposer sweeps store aggregate episode token means and evaluate archive eligibility under $\mathcal{E}$; during online search, intermediate parent selection also referenced the aggregate $\mathcal{E}$ flag, with all $24/24$ intermediate parent choices and all $6/6$ final selections on GLM-5.3 measure-then-revise matching between $\mathcal{E}_{\mathcal{S}}$ and $\mathcal{E}$ ($20/24$ intermediate parents match under discard-and-resample).

Across the 75 full-loop runs, 66 evaluate 13 harnesses and the nine 400-word prose runs evaluate 12 ($66 \times 13 + 9 \times 12 = 966$ harnesses; $492{,}143$ conversations total including draft screening, allocator baselines, and re-scoring).

\begin{table}[!ht]
\centering
\caption{Complete experiment ledger across all evaluated settings.}
\label{tab:full_ledger}
\vspace{3pt}
\resizebox{\textwidth}{!}{%
\begin{tabular}{@{}lllrrrrrrr@{}}
\toprule
Paper Label & Results File (\texttt{experiments/results/}) & Split & Seeds & $\Delta_{\mathrm{ship}}$ & $t_{\mathrm{task}}$ & $t_{\mathrm{pool}}$ & $\rho_{\mathrm{start}}$ & Over Cap & $\Delta_{\mathrm{all}}^\star$ / $\Delta_{\mathrm{elig}}^\star$ \\
\midrule
400w prose / transcript$^\S$ & \texttt{prose\_400w\_transcript.md} & train$\to$test & 3 (9) & \pmsd{-0.006}{0.006} & $-0.44$ & $-0.50$ & $6/33$ & $0/33$ & $0.003$ / $0.003$ \\
150w prose / transcript$^\S$ & \texttt{prose\_150w\_transcript.md} & train$\to$test & 3 (6) & \pmsd{-0.030}{0.004} & $-1.57$ & $-1.80$ & $3/36$ & $0/36$ & $0.013$ / $0.013$ \\
150w prose / diffs & \texttt{prose\_150w\_diff\_train\_test.md} & train$\to$test & 3 & \pmsd{-0.003}{0.011} & $-0.15$ & $-0.16$ & $14/36$ & $0/36$ & $0.036$ / $0.036$ \\
Prose + guard / diffs & \texttt{prose\_guard\_diff.md} & train$\to$test & 3 & \pmsd{-0.007}{0.021} & $-0.53$ & $-0.55$ & $3/36$ & $0/36$ & $0.003$ / $0.003$ \\
150w prose / diffs ($50/65$) & \texttt{prose\_150w\_diff\_retail.md} & $50/65$ test & 3 & \pmsd{0.000}{0.049} & $0.00$ & $0.00$ & $10/36$ & $0/36$ & $0.024$ / $0.024$ \\
150w prose / contrast & \texttt{prose\_150w\_contrast.md} & $50/65$ test & 3 & \pmsd{0.002}{0.016} & $0.11$ & $0.09$ & $18/36$ & $0/36$ & $0.060$ / $0.060$ \\
\midrule
GPT-OSS-120B; no cap control & \texttt{gpt\_oss\_no\_cap.md} & $50/65$ test & 3 & \pmsd{0.034}{0.064} & $1.86$ & $1.94$ & $28/36$ & $27/36$ & $0.096$ / $0.056$ \\
GPT-OSS-120B; history + discard & \texttt{gpt\_oss\_discard.md} & $50/65$ test & 3 & \pmsd{0.048}{0.013} & $2.05$ & $2.34$ & $20/36$ & $8/36$ & $0.070$ / $0.053$ \\
Gemini 4 Argon & \texttt{gemini4\_argon\_discard.md} & $50/65$ test & 3 & \pmsd{0.077}{0.060} & $3.33$ & $3.32$ & $34/36$ & $21/36$ & $0.125$ / $0.111$ \\
Gemini 3.8 Flash & \texttt{gemini38\_flash\_discard.md} & $50/65$ test & 3 & \pmsd{0.091}{0.056} & $3.88$ & $4.25$ & $36/36$ & $26/36$ & $0.161$ / $0.125$ \\
DeepSeek V4.1 Flash & \texttt{deepseek\_v41\_discard.md} & $50/65$ test & 3 & \pmsd{0.079}{0.024} & $2.83$ & $3.71$ & $36/36$ & $21/36$ & $0.138$ / $0.109$ \\
GLM-5.2; history + discard & \texttt{glm52\_discard.md} & $50/65$ test & 3 & \pmsd{0.092}{0.062} & $3.39$ & $3.67$ & $34/36$ & $19/36$ & $0.138$ / $0.123$ \\
GLM-5.2; history only & \texttt{glm52\_history\_only.md} & $50/65$ test & 3 & \pmsd{0.046}{0.040} & $1.66$ & $1.92$ & $34/36$ & $29/36$ & $0.137$ / $0.075$ \\
GLM-5.2; discard only & \texttt{glm52\_discard\_only.md} & $50/65$ test & 3 & \pmsd{0.055}{0.031} & $2.19$ & $2.57$ & $36/36$ & $27/36$ & $0.132$ / $0.070$ \\
GLM-5.2; airline & \texttt{glm52\_airline\_discard.md} & $20/30$ air. & 3 & \pmsd{0.052}{0.051} & $1.23$ & $1.62$ & $34/36$ & $27/36$ & $0.152$ / $0.081$ \\
GLM-5.3; discard (search cap$^\star$) & \texttt{glm53\_discard.md} & $50/65$ test & 6 & \pmsd{0.054}{0.064} & $3.10$ & $3.59$ & $63/72$ & $38/72$ & $0.113$ / $0.097$ \\
GLM-5.3; discard (combined cap) & \texttt{glm53\_discard.md} & $50/65$ test & 6 & \pmsd{0.063}{0.071} & $3.69$ & $4.27$ & $63/72$ & $43/72$ & $0.113$ / $0.090$ \\
GLM-5.3; revise (search cap$^\star$) & \texttt{glm53\_revise.md} & $50/65$ test & 6 & \pmsd{0.116}{0.029} & $5.74$ & $7.59$ & $68/72$ & $2/72$ & $0.123$ / $0.117$ \\
GLM-5.3; revise (combined cap) & \texttt{glm53\_revise.md} & $50/65$ test & 6 & \pmsd{0.116}{0.029} & $5.74$ & $7.59$ & $68/72$ & $4/72$ & $0.123$ / $0.117$ \\
DeepSeek V4.1 Flash; history + revise & \texttt{deepseek\_v41\_revise.md} & $50/65$ test & 3 & \pmsd{0.074}{0.036} & $2.70$ & $3.18$ & $28/36$ & $0/36$ & $0.118$ / $0.118$ \\
\midrule
Uniform \texttt{medium} reference & \texttt{uniform\_medium.md} & $65$ held-out & 3 & \pmsd{0.091}{0.021} & $3.23$ & $4.00$ & -- & -- & -- \\
From \texttt{medium}; $1.2\times$ cap & \texttt{medium\_cap\_1p2x.md} & $50/65$ test & 3 & \pmsd{-0.012}{0.021} & $-1.04$ & $-1.04$ & $2/36$ & $12/36$ & $0.031$ / $0.000$ \\
From \texttt{medium}; $2.0\times$ cap & \texttt{medium\_cap\_2x.md} & $50/65$ test & 3 & \pmsd{-0.050}{0.078} & $-2.67$ & $-2.61$ & $6/36$ & $0/36$ & $0.009$ / $0.009$ \\
\midrule
Confidence-Turn ($q{=}0.20$) & \texttt{learned\_allocators.md} & $50/65$ test & 3 & \pmsd{0.026}{0.021} & -- & -- & -- & $0/3$ & -- \\
Confidence-Turn ($q{=}0.425$, post hoc) & \texttt{learned\_allocators.md} & $50/65$ test & 3 & \pmsd{0.063}{0.038} & -- & -- & -- & $0/3$ & -- \\
BAAR-Turn (ModernBERT) & \texttt{learned\_allocators.md} & $50/65$ test & 3 & \pmsd{0.065}{0.012} & -- & -- & -- & $1/3$ & -- \\
CPO-RL-Turn (primal-dual) & \texttt{learned\_allocators.md} & $50/65$ test & 3 & \pmsd{0.068}{0.047} & -- & -- & -- & $0/3$ & -- \\
\bottomrule
\end{tabular}%
}\\[2pt]
{\raggedright\scriptsize Gains and root-inclusive oracles in decimal reward units. $^\star$Search-only eligibility ($\hat{C}_{\mathcal{S}} \le 2\hat{C}_{\mathcal{S}}(h_0)$); GLM-5.3 ``combined cap'' rows evaluate the same 6 seeds retrospectively under $C(h) \le 2C(h_0)$. $^\S$Cohort totals in parentheses include reward-agnostic/blind control seeds ($9 \times 12 = 108$ harnesses in \texttt{runs\_idea1}; $6 \times 13 = 78$ in \texttt{runs\_idea1b}; $75$ loop runs and $966$ harnesses overall). $t_{\mathrm{task}}$: task-paired ($n{=}65$ retail, $30$ airline, $115$ train$\to$test); $t_{\mathrm{pool}}$: task$\times$seed pairs ($195$ for 3 seeds, $390$ for 6 seeds). For from-\texttt{medium} controls, candidate-only $\Delta_{\mathrm{cand}}^\star$ is $0.024 / -0.043$ ($1.2\times$) and $-0.010 / -0.010$ ($2.0\times$).\par}
\end{table}

\FloatBarrier
\section{Router Interface, Revision Prompt, and Case Study}
\label{app:prompts}

This appendix describes the compute router interface, the revision prompt used during budget calibration, and a representative example of how measure-then-revise modifies an expensive router while preserving its reasoning strategy.

\subsection{Compute Router Interface}

The compute router controls the reasoning effort of the executor at two stages: before response generation and before tool execution. \Cref{tab:interface} summarizes the available reasoning levels, router inputs, return values, and validation requirements.

\begin{table}[!ht]
\centering
\caption{Complete two-stage compute router interface, call-stage eligibility rules, and static AST validation on \texttt{gpt-oss-120b} (vLLM).}
\label{tab:interface}
\vspace{3pt}
\small
\begin{tabular}{@{}lp{12.8cm}@{}}
\toprule
Component & Specification \\
\midrule
Effort tiers & \texttt{reasoning\_effort} $\in \{\texttt{low}, \texttt{medium}, \texttt{high}\}$ per call (temp $0$, \texttt{max\_tokens} $4{,}096$). Root runs all-\texttt{low} ($\sim\!1.65\text{k}$ mean output tokens/episode; uniform \texttt{medium} $1.64\times$, uniform \texttt{high} $4.4\times$). \\
\texttt{turn} stage & Invoked before generation with \texttt{info} (\texttt{step}, \texttt{last\_role}, \texttt{last\_tool}, \texttt{last\_content[:2000]}, \texttt{writes\_done}, \texttt{errors\_seen}, \texttt{n\_tool\_calls}) and mutable dict \texttt{ctx}; returns a tier string or \texttt{None} (keeping current default). \\
\texttt{call} stage & Invoked only when a \texttt{low} or \texttt{medium} turn emits a tool call (\texttt{name}, \texttt{args}, \texttt{effort} in \texttt{info}). Returning a strictly higher tier regenerates the turn once before tool execution; \emph{both} the discarded attempt and the regenerated call count toward episode output tokens. \\
Validation & $\le 150$ lines, valid Python AST and \texttt{route(stage, info, ctx)} signature; runtime exceptions fall back to the active tier. Routers have zero access to environment state, tool implementations, domain policy text, or evaluators. \\
\bottomrule
\end{tabular}
\end{table}

At the turn stage, the router receives information about the current conversation and assigns low, medium, or high reasoning effort. At the call stage, the router can request one higher-effort regeneration when a low- or medium-effort response contains a tool call.

Both the original and regenerated responses contribute to the episode's output-token consumption. Router code is limited to 150 lines and is checked for valid Python syntax and function signature before evaluation. If a router raises an exception during execution, the executor retains its current reasoning effort.

\subsection{Exact Revision Prompt}

When a candidate router exceeds the token budget after the first rollout (50 search episodes), or exceeds $95\%$ of the budget after full search evaluation (150 episodes), the loop constructs a revision prompt using measured costs from search episodes only.

The prompt includes the current router code, mean output tokens per conversation, token consumption and turn counts at each reasoning level, and the number of higher-effort regenerations.

The proposer is instructed to reduce token consumption toward $85\%$ of the budget while preserving the router's original reasoning strategy.

The exact prompt used in our experiments is shown below.

\begin{quote}\small\raggedright\ttfamily
You wrote the compute router below for a customer-service agent. It was just measured and it is OVER the token budget, so it cannot be selected. Revise it so that it fits the budget while keeping as much of its benefit as possible.\\
\{ROUTE\_SPEC\}\\
\# Your router\\
\{guard\}\\
\# Measured cost of this router (\{n\_eps\} conversations)\\
- output tokens per conversation: \{tok:.0f\} = \{ratio:.2f\}x the budget of \{cap:.0f\}. Target: at most \{target:.0f\} tokens per conversation (0.85x the budget, to leave a margin).\\
- per conversation, by effort: low: \{turns\_low\} turns, \{tok\_low\} tokens; medium: \{turns\_med\} turns, \{tok\_med\} tokens; high: \{turns\_high\} turns, \{tok\_high\} tokens\\
- turns regenerated at a higher effort by the "call" stage (the cheaper attempt is paid for and discarded): \{regen:.1f\} per conversation\\
Cut cost where it buys least: fewer turns at "high" (use "medium"), fewer regenerations, effort only on the turn types your router targets. Keep the idea of the router; do not fall back to all-low.
\end{quote}

\subsection{Case Study: Revising an Over-Budget Router}

We illustrate the measure-then-revise process using Candidate 10 from GLM-5.3 Seed 2, which was ultimately selected by the harness evolution loop and achieved a held-out improvement of $12.3$\,pp.

During the initial screening of 50 search conversations, the router consumes an average of $4{,}543$ output tokens per conversation, or $2.76\times$ the root harness's mean of $1{,}645$ tokens. This exceeds the budget of $3{,}289$ tokens.

After receiving measured token consumption at each reasoning level, the proposer revises the router in a single attempt. The revised router makes four principal changes:
\begin{enumerate}[leftmargin=18pt,itemsep=2pt,parsep=0pt,topsep=3pt]
    \item It replaces high-effort regeneration of write operations with medium-effort regeneration, and only regenerates responses initially produced at low effort.
    \item It reserves high reasoning effort for requests involving three or more distinct actions, rather than using high effort for all multi-part requests.
    \item It reduces reasoning effort after routine tool responses, including successful write operations and product catalog listings.
    \item It preserves medium reasoning effort when interpreting order details, comparing product variants, and responding to tool errors.
\end{enumerate}

Across its 345 evaluated conversations, the revised router consumes an average of $2{,}866$ output tokens per conversation, corresponding to $1.74\times$ the root harness's mean token consumption. Its execution includes $2{,}597$ low-effort turns, $2{,}395$ medium-effort turns, $80$ high-effort turns, and $294$ regenerations.

The revised router satisfies the token budget and achieves a held-out gain of $12.3$\,pp over the root harness.

The following unified diff shows the modifications between the initial draft and the revised router.

\begin{tcolorbox}[enhanced, breakable, colback=white, colframe=cardborder, boxrule=0.6pt, arc=5pt, left=8pt, right=8pt, top=7pt, bottom=8pt, before skip=6pt, after skip=8pt]
{\footnotesize\sffamily\color{codelabel}Unified diff: over-budget initial draft (\texttt{-}) vs.\ single-pass revised router (\texttt{+})}\par\vspace{5pt}
{\setlength{\fboxsep}{1.3pt}%
\difffileold{-{}-{}-\ draft\_slot10.py\ (4,543\ tok/ep\ =\ 2.76x\ root;\ OVER\ CAP)}
\difffilenew{+++\ revised\_node10.py\ (2,866\ tok/ep\ =\ 1.74x\ root;\ SHIPPED\ 0.123)}
\diffhunk{@@\ -2,8\ +2,22\ @@}
\diffctx{\ import\ re}
\diffctx{\ }
\diffrem{-\#\ Database-writing\ tools:\ every\ attempt\ to\ execute\ one\ is\ regenerated\ at\ high}
\diffrem{-\#\ effort\ so\ the\ agent\ re-checks\ order\ ids,\ item\ ids,\ variants,\ payment\ methods,}
\diffrem{-\#\ ordering\ of\ operations\ (e.g.\ address\ before\ item\ modification),\ and\ whether}
\diffrem{-\#\ the\ user\ actually\ requested/confirmed\ this\ exact\ write.}
\diffadd{+\#\ Compute\ router,\ revised\ for\ the\ 2.0x\ output-token\ budget.}
\diffadd{+\#\ Cost\ changes\ vs.\ the\ previous\ version\ (measured\ 4543\ tok/conv,\ budget\ share}
\diffadd{+\#\ 2795):}
\diffadd{+\#\ \ \ *\ No\ more\ blanket\ HIGH:\ high\ turns\ cost\ \textasciitilde{}673\ tokens\ each\ and\ the\ budget}
\diffadd{+\#\ \ \ \ \ report\ shows\ they\ are\ where\ the\ extra\ care\ bought\ the\ least.\ HIGH\ is\ now}
\diffadd{+\#\ \ \ \ \ reserved\ for\ the\ rare\ 3+-action\ mega-requests\ only.}
\diffadd{+\#\ \ \ *\ Write\ calls\ are\ regenerated\ at\ MEDIUM\ (not\ high)\ and\ only\ when\ the}
\diffadd{+\#\ \ \ \ \ producing\ turn\ was\ low:\ a\ medium\ attempt\ is\ not\ re-generated,\ which}
\diffadd{+\#\ \ \ \ \ removes\ the\ double-pay\ on\ the\ 1.7\ regenerations/conversation.}
\diffadd{+\#\ \ \ *\ Turns\ after\ writes\ and\ after\ list\_all\_product\_types\ drop\ to\ low:\ the}
\diffadd{+\#\ \ \ \ \ write\ itself\ is\ already\ guarded\ at\ the\ \textquotedbl{}call\textquotedbl{}\ stage,\ and\ the\ tool\ result}
\diffadd{+\#\ \ \ \ \ already\ contains\ the\ state\ to\ report.\ Only\ get\_order\_details\ /}
\diffadd{+\#\ \ \ \ \ get\_product\_details\ results\ (picking\ the\ right\ order/item/variant)\ stay}
\diffadd{+\#\ \ \ \ \ at\ medium,\ plus\ anything\ after\ a\ tool\ error.}
\diffadd{+}
\diffadd{+\#\ Database-writing\ tools:\ an\ attempt\ from\ a\ low-effort\ turn\ is\ regenerated\ at}
\diffadd{+\#\ medium\ effort\ so\ the\ agent\ re-checks\ order\ ids,\ item\ ids,\ variants,\ payment}
\diffadd{+\#\ methods,\ and\ whether\ the\ user\ actually\ requested\ this\ exact\ write.}
\diffctx{\ WRITE\_TOOLS\ =\ \{}
\diffctx{\ \ \ \ \ \textquotedbl{}cancel\_pending\_order\textquotedbl{},}
\diffhunk{@@\ -39,9\ +53,7\ @@}
\diffctx{\ \ \ \ \ r\textquotedbl{}please\ do|go\ ahead|proceed)\textbackslash{}b[\textbackslash{}s.!,:)]*(i(\textquotesingle{}|\ a)?m\ all\ set.*)?\$\textquotedbl{},\ re.I)}
\diffctx{\ }
\diffrem{-TURNS\_MEDIUM\_TOOLS\ =\ \{}
\diffrem{-\ \ \ \ \#\ Turns\ reacting\ to\ these\ results\ need\ care:\ correct\ item/order\ pickup,}
\diffrem{-\ \ \ \ \#\ variant\ comparison,\ summarising\ the\ new\ database\ state\ faithfully.}
\diffrem{-\ \ \ \ \textquotedbl{}get\_order\_details\textquotedbl{},\ \textquotedbl{}get\_product\_details\textquotedbl{},\ \textquotedbl{}list\_all\_product\_types\textquotedbl{},}
\diffrem{-\}}
\diffadd{+\#\ Results\ that\ still\ need\ care\ to\ read:\ correct\ order/item\ pickup\ and\ variant}
\diffadd{+\#\ comparison.\ list\_all\_product\_types\ results\ dropped\ to\ low\ (plain\ list).}
\diffadd{+CAREFUL\_TOOLS\ =\ \{\textquotedbl{}get\_order\_details\textquotedbl{},\ \textquotedbl{}get\_product\_details\textquotedbl{}\}}
\diffctx{\ }
\diffctx{\ }
\diffhunk{@@\ -50,8\ +62,10\ @@}
\diffctx{\ \ \ \ \ effort\ =\ info.get(\textquotedbl{}effort\textquotedbl{})}
\diffctx{\ \ \ \ \ if\ name\ in\ WRITE\_TOOLS\ or\ name\ ==\ \textquotedbl{}transfer\_to\_human\_agents\textquotedbl{}:}
\diffrem{-\ \ \ \ \ \ \ \ \#\ Premature/incorrect\ transfers\ and\ wrong-argument\ writes\ are\ the\ main}
\diffrem{-\ \ \ \ \ \ \ \ \#\ failure\ modes;\ regenerate\ the\ whole\ turn\ at\ high\ effort.}
\diffrem{-\ \ \ \ \ \ \ \ if\ effort\ !=\ \textquotedbl{}high\textquotedbl{}:}
\diffrem{-\ \ \ \ \ \ \ \ \ \ \ \ return\ \textquotedbl{}high\textquotedbl{}}
\diffadd{+\ \ \ \ \ \ \ \ \#\ Regenerate\ only\ cheap\ (low)\ attempts,\ and\ only\ up\ to\ medium:\ this}
\diffadd{+\ \ \ \ \ \ \ \ \#\ re-checks\ arguments\ and\ premature-transfer\ conditions\ without\ paying}
\diffadd{+\ \ \ \ \ \ \ \ \#\ \textasciitilde{}673\ tokens\ for\ a\ high\ turn\ or\ discarding\ an\ already-careful\ medium}
\diffadd{+\ \ \ \ \ \ \ \ \#\ attempt,\ which\ is\ what\ blew\ the\ budget\ before.}
\diffadd{+\ \ \ \ \ \ \ \ if\ effort\ ==\ \textquotedbl{}low\textquotedbl{}:}
\diffadd{+\ \ \ \ \ \ \ \ \ \ \ \ return\ \textquotedbl{}medium\textquotedbl{}}
\diffctx{\ \ \ \ \ \ \ \ \ return\ None}
\diffctx{\ \ \ \ \ \#\ Reads\ are\ cheap;\ only\ bump\ if\ the\ model\ has\ been\ stumbling.}
\diffhunk{@@\ -72,16\ +86,14\ @@}
\diffctx{\ \ \ \ \ \ \ \ \ actions\ =\ [k\ for\ k,\ pat\ in\ ACTION\_PATTERNS.items()\ if\ pat.search(last\_content)]}
\diffctx{\ \ \ \ \ \ \ \ \ has\_q\ =\ bool(QUESTION\_RE.search(last\_content))}
\diffrem{-\ \ \ \ \ \ \ \ \#\ Multi-part\ requests\ (several\ actions,\ or\ question\ +\ action)\ are\ the}
\diffrem{-\ \ \ \ \ \ \ \ \#\ hardest\ turns:\ parts\ get\ dropped,\ wrong\ variants\ picked,\ premature}
\diffrem{-\ \ \ \ \ \ \ \ \#\ transfers\ happen\ here.\ Spend\ high\ effort.}
\diffrem{-\ \ \ \ \ \ \ \ if\ len(actions)\ >=\ 2\ or\ (has\_q\ and\ actions):}
\diffadd{+\ \ \ \ \ \ \ \ \#\ Genuinely\ extreme\ multi-part\ requests\ (three\ or\ more\ distinct}
\diffadd{+\ \ \ \ \ \ \ \ \#\ actions)\ are\ the\ one\ case\ still\ worth\ high\ effort;\ they\ are\ rare,}
\diffadd{+\ \ \ \ \ \ \ \ \#\ so\ the\ cost\ stays\ small.}
\diffadd{+\ \ \ \ \ \ \ \ if\ len(actions)\ >=\ 3:}
\diffctx{\ \ \ \ \ \ \ \ \ \ \ \ \ return\ \textquotedbl{}high\textquotedbl{}}
\diffrem{-\ \ \ \ \ \ \ \ \#\ Single\ action:\ needs\ correct\ detail\ gathering\ (ids,\ reasons,}
\diffrem{-\ \ \ \ \ \ \ \ \#\ payment\ methods)\ but\ is\ routine\ ->\ medium.}
\diffrem{-\ \ \ \ \ \ \ \ if\ actions:}
\diffrem{-\ \ \ \ \ \ \ \ \ \ \ \ return\ \textquotedbl{}medium\textquotedbl{}}
\diffrem{-\ \ \ \ \ \ \ \ \#\ Pure\ catalogue/information\ question:\ needs\ the\ right\ product\ lookup}
\diffrem{-\ \ \ \ \ \ \ \ \#\ and\ an\ accurate\ answer\ ->\ medium.}
\diffrem{-\ \ \ \ \ \ \ \ if\ has\_q:}
\diffadd{+\ \ \ \ \ \ \ \ \#\ Multi-part\ (2\ actions,\ or\ question\ +\ action)\ and\ single-action\ /}
\diffadd{+\ \ \ \ \ \ \ \ \#\ single-question\ turns:\ medium\ is\ enough\ to\ gather\ ids,\ reasons\ and}
\diffadd{+\ \ \ \ \ \ \ \ \#\ payment\ methods\ and\ to\ avoid\ dropping\ parts,\ at\ \textasciitilde{}255\ tokens\ a\ turn}
\diffadd{+\ \ \ \ \ \ \ \ \#\ instead\ of\ \textasciitilde{}673.}
\diffadd{+\ \ \ \ \ \ \ \ if\ len(actions)\ >=\ 2\ or\ (has\_q\ and\ actions)\ or\ actions\ or\ has\_q:}
\diffctx{\ \ \ \ \ \ \ \ \ \ \ \ \ return\ \textquotedbl{}medium\textquotedbl{}}
\diffctx{\ \ \ \ \ \ \ \ \ \#\ Greetings,\ identity\ info,\ chit-chat,\ bare\ confirmations\ ->\ low.}
\diffhunk{@@\ -89,14\ +101,14\ @@}
\diffctx{\ }
\diffctx{\ \ \ \ \ if\ last\_role\ ==\ \textquotedbl{}tool\textquotedbl{}:}
\diffrem{-\ \ \ \ \ \ \ \ if\ last\_tool\ in\ WRITE\_TOOLS:}
\diffrem{-\ \ \ \ \ \ \ \ \ \ \ \ \#\ Must\ report\ the\ resulting\ database\ state\ exactly.}
\diffadd{+\ \ \ \ \ \ \ \ if\ errors\ >\ 0:}
\diffctx{\ \ \ \ \ \ \ \ \ \ \ \ \ return\ \textquotedbl{}medium\textquotedbl{}}
\diffrem{-\ \ \ \ \ \ \ \ if\ last\_tool\ in\ TURNS\_MEDIUM\_TOOLS:}
\diffadd{+\ \ \ \ \ \ \ \ if\ last\_tool\ in\ CAREFUL\_TOOLS:}
\diffctx{\ \ \ \ \ \ \ \ \ \ \ \ \ \#\ Picking\ the\ right\ order/item/variant\ from\ these\ results\ is\ where}
\diffctx{\ \ \ \ \ \ \ \ \ \ \ \ \ \#\ low\ effort\ makes\ mistakes;\ also\ prevents\ giving\ up\ on\ catalog}
\diffctx{\ \ \ \ \ \ \ \ \ \ \ \ \ \#\ data\ instead\ of\ calling\ get\_product\_details.}
\diffctx{\ \ \ \ \ \ \ \ \ \ \ \ \ return\ \textquotedbl{}medium\textquotedbl{}}
\diffrem{-\ \ \ \ \ \ \ \ if\ errors\ >\ 0:}
\diffrem{-\ \ \ \ \ \ \ \ \ \ \ \ return\ \textquotedbl{}medium\textquotedbl{}}
\diffadd{+\ \ \ \ \ \ \ \ \#\ Write\ results\ and\ plain\ listings:\ the\ risky\ decision\ already}
\diffadd{+\ \ \ \ \ \ \ \ \#\ happened\ (guarded\ at\ the\ \textquotedbl{}call\textquotedbl{}\ stage)\ and\ the\ result\ text\ contains}
\diffadd{+\ \ \ \ \ \ \ \ \#\ the\ state\ to\ report,\ so\ a\ low\ turn\ is\ enough.}
\diffctx{\ \ \ \ \ \ \ \ \ return\ \textquotedbl{}low\textquotedbl{}}
}
\end{tcolorbox}

This example illustrates how measured runtime costs can guide targeted changes to an existing compute router. Rather than uniformly reducing reasoning effort, the revised router preserves medium-effort reasoning for selected operations while restricting more expensive high-effort generations.

\section{Additional Experimental Results and Controls}
\label{app:extended}

\begin{figure}[!ht]
\centering
\includegraphics[width=0.50\textwidth]{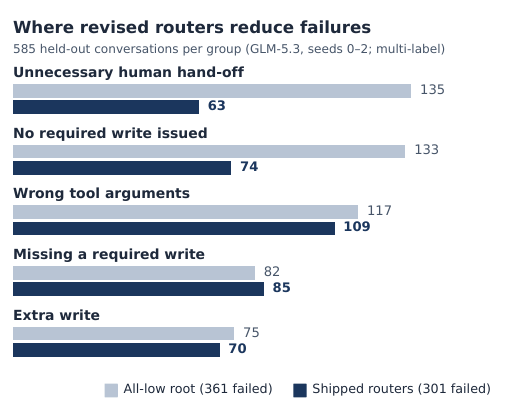}
\vspace{-8pt}
\caption{Failure-tag counts on held-out tasks (GLM-5.3, seeds 0--2, 585 conversations/group). Multi-label comparison between all-\texttt{low} roots and shipped revised routers (\cref{tab:failures}).}
\label{fig:failures}
\vspace{-4pt}
\end{figure}

\Cref{tab:telemetry,tab:failures} provide the numerical results underlying \cref{fig:allocation,fig:failures}. \Cref{tab:telemetry} reports reasoning effort allocation, token consumption, and regeneration frequency across different router groups. \Cref{tab:failures} reports the frequency of different failure types on held-out tasks.

These analyses are descriptive and are based on the first three GLM-5.3 seeds.

\begin{table}[!ht]
\centering
\caption{Per-conversation search effort allocation across GLM-5.3 router groups (seeds 0--2).}
\label{tab:telemetry}
\vspace{3pt}
\small
\begin{tabular}{@{}lrrrrrrr@{}}
\toprule
Router Group ($n$) & Low Turns & Med Turns & High Turns & Tok$_{\mathrm{M}}$ & Tok$_{\mathrm{H}}$ & Regen / Conv & Mean Out Tok \\
\midrule
All-low root ($3$) & $13.1$ & $0.0$ & $0.0$ & $0$ & $0$ & $0.00$ & $1{,}673$ \\
Revise: initial drafts ($33$) & $5.3$ & $6.7$ & $2.7$ & $1{,}454$ & $1{,}821$ & $1.10$ & $4{,}019$ \\
Revise: revised routers ($33$) & $6.5$ & $6.8$ & $0.8$ & $1{,}479$ & $541$ & $0.63$ & $2{,}915$ \\
Discard: survivors ($14$) & $9.2$ & $3.8$ & $1.5$ & $841$ & $977$ & $1.31$ & $3{,}028$ \\
Discard: over-cap ($22$) & $5.9$ & $6.3$ & $2.1$ & $1{,}332$ & $1{,}500$ & $0.86$ & $3{,}651$ \\
\bottomrule
\end{tabular}\\[2pt]
{\raggedright\scriptsize Underlying values for \cref{fig:allocation} ($50$ search episodes for Rollout-1 drafts, $150$ for full-search drafts and final candidates; descriptive). Per-tier tokens omit small unattributed overhead included in total output tokens.\par}
\end{table}

\begin{table}[!ht]
\centering
\caption{Automatic failure-tag counts on GLM-5.3 held-out episodes (seeds 0--2, $585$ conversations/group).}
\label{tab:failures}
\vspace{3pt}
\small
\begin{tabular}{@{}lrr@{}}
\toprule
Held-Out Outcome / Failure Tag & Root (All-\texttt{low}) & Shipped (Candidates 12, 5, 10) \\
\midrule
Success rate & $0.383$ ($38.3\%$) & $0.485$ ($48.5\%$) \\
Failed conversations (of $585$) & $361$ & $301$ \\
\midrule
Unneeded hand-off to a human & $135$ & $63$ \\
No write at all (one required) & $133$ & $74$ \\
Right tool, wrong arguments & $117$ & $109$ \\
Missing a required write & $82$ & $85$ \\
Extra write & $75$ & $70$ \\
\midrule
Conversations without \texttt{get\_user\_details} (success rate there) & $150$ ($0.240$) & $95$ ($0.389$) \\
\bottomrule
\end{tabular}\\[2pt]
{\raggedright\scriptsize Underlying values for \cref{fig:failures}. Multi-label ground-truth write comparisons (descriptive; H100 runs only).\par}
\end{table}

\FloatBarrier
\subsection{Alternative Candidate Selection Rules}

We evaluate six alternative candidate ranking rules using the recorded search episodes from 12 independent runs across four proposer models. Each candidate is evaluated using 150 search episodes.

The standard selection rule ranks candidates by their mean search performance and achieves a held-out gain of $8.5$\,pp. Combining candidate and parent scores, with weights of $0.7$ and $0.3$ respectively, reduces the gain to $8.1$\,pp. Ranking candidates using a task-paired lower confidence bound achieves $8.5$\,pp, while ranking by the minimum performance across three rollouts achieves $7.8$\,pp.

Two alternative selection rules produce higher observed gains. Penalizing candidates by $0.05$ for each unit of the root-normalized token multiplier achieves $9.2$\,pp. Ranking candidates by the median of their three rollout scores achieves $10.0$\,pp and selects the best eligible router in three runs: Gemini 4 Argon Seed 0, Gemini 3.8 Flash Seed 0, and GLM-5.2 Seed 0.

These results suggest that alternative ranking rules may improve candidate selection. However, because the analysis is conducted retrospectively using existing evaluation records, further independent experiments are needed to establish whether these improvements generalize.

\subsection{Candidate Recovery, Search-Round Dynamics, and Gate Sensitivity}
\label{app:sensitivity}

\begin{figure}[!ht]
\centering
\includegraphics[width=0.52\textwidth]{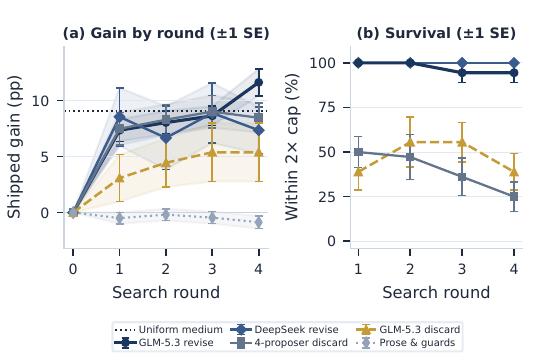}
\vspace{-8pt}
\caption{Progress across search rounds and cap survival ($\pm 1\text{ SE}$ across seeds). (a) Shipped held-out gain across Rounds 0--4. (b) Share of candidates within the $2\times$ cap per round.}
\label{fig:generations}
\vspace{-4pt}
\end{figure}

\Cref{fig:generations} tracks shipped held-out gain and within-budget candidate survival across Rounds 0--4, illustrating how stronger proposers capture most held-out accuracy gains in Round 1 while subsequent rounds determine whether candidates remain within the $2\times$ token budget.

We analyze the generation and evaluation history of 144 candidates across the four Vertex AI open-loop proposer settings.

Only 22 of 144 initial drafts ($15.3\%$) satisfy the search token budget. Of these, 18 remain within the budget after full evaluation. Discard-and-resample recovers an additional 39 candidates, increasing the total number of budget-eligible candidates to 57 of 144 ($39.6\%$).

Nine of the 12 selected routers originate from replacement drafts. Including these candidates increases the selected harnesses' mean held-out gain from $5.0$ to $8.5$\,pp.

Across all 66 budget-eligible retail candidates included in the repair-bias analysis, the 45 candidates recovered through replacement achieve a mean held-out gain of \pmsd{7.5}{4.8}\,pp. The 21 candidates that satisfy the budget on their first attempt achieve \pmsd{7.1}{3.3}\,pp. The correlations between search and held-out performance are $0.10$ and $-0.25$, respectively.

The observed mean gains are similar, suggesting that recovering initially expensive candidates does not necessarily reduce their held-out performance.

\Cref{tab:gate_sensitivity} presents a retrospective sensitivity analysis of two measure-then-revise parameters: the maximum number of revision attempts, $K$, and the full-search revision threshold, $\gamma_{\mathrm{trig}}$.

For the revision-limit analysis, we examine recorded draft histories and determine what the budget violation rate would have been if the revision process had stopped earlier; on GLM-5.3, combined-cost violations decrease from $86.1\%$ ($K = 0$) to $31.9\%$, $9.7\%$, and $5.6\%$ across $K \in \{1, 2, 3\}$ ($83.3\%$ to $0.0\%$ on DeepSeek). For intermediate drafts without held-out evaluations, budget status is determined using measured search costs.

For the revision-threshold analysis, we examine all 144 evaluated GLM-5.3 routers and measure how frequently candidates that satisfy a given search threshold subsequently exceed the $2\times$ combined-cost budget: applying $\gamma_{\mathrm{trig}} = 0.95B$ reduces post-search budget drift from $11.9\%$ ($\gamma_{\mathrm{trig}} = 1.00B$) to $3.3\%$, while the single-rollout screen achieves $92.7\%$ precision ($38/41$) relative to the full-search trigger.

These analyses quantify the trade-off between revision overhead, candidate eligibility, and the risk of exceeding the budget during full evaluation.

\begin{table}[!ht]
\centering
\caption{Measure-then-revise gate sensitivity across (a) revision limit $K$ and (b) full-search trigger $\gamma_{\mathrm{trig}}$.}
\label{tab:gate_sensitivity}
\vspace{3pt}
{\small
\textbf{(a) Revision limit $K$ across recorded draft chains}\\[3pt]
\begin{tabular}{@{}lrrr@{}}
\toprule
Revision Limit $K$ & GLM-5.3 Over Cap & DeepSeek Over Cap & GLM-5.3 Revision Eps \\
\midrule
$K = 0$ (no revision) & $62/72$ ($86.1\%$) & $30/36$ ($83.3\%$) & $50$ \\
$K = 1$ & $23/72$ ($31.9\%$) & $6/36$ ($16.7\%$) & $3{,}950$ \\
$K = 2$ & $7/72$ ($9.7\%$) & $1/36$ ($2.8\%$) & $6{,}000$ \\
$K = 3$ (default) & $4/72$ ($5.6\%$) & $0/36$ ($0.0\%$) & $6{,}300$ \\
\bottomrule
\end{tabular}

\vspace{8pt}
\textbf{(b) Full-search trigger $\gamma_{\mathrm{trig}}$ across all $144$ evaluated GLM-5.3 routers}\\[3pt]
\begin{tabular}{@{}lrr@{}}
\toprule
Trigger Threshold $\gamma_{\mathrm{trig}}$ & Search-Passing Routers & All-Episode Drift ($>2\times$) \\
\midrule
$\gamma_{\mathrm{trig}} = 0.90$ ($\le 1.80\times$ root) & $53/144$ & $0/53$ ($0.0\%$) \\
$\gamma_{\mathrm{trig}} = 0.95$ ($\le 1.90\times$, ours) & $90/144$ & $3/90$ ($3.3\%$) \\
$\gamma_{\mathrm{trig}} = 0.98$ ($\le 1.96\times$ root) & $96/144$ & $4/96$ ($4.2\%$) \\
$\gamma_{\mathrm{trig}} = 1.00$ ($\le 2.00\times$, unbuffered) & $109/144$ & $13/109$ ($11.9\%$) \\
\bottomrule
\end{tabular}
}\\[3pt]
{\raggedright\scriptsize In (a), counterfactual stopping at $K < n_{\mathrm{trims}}$ evaluates cap status from draft $K$'s measured search cost; at $K = 0$, $50$ episodes reflects one draft retained when revision returned empty. In (b), post-search drift pools all $144$ evaluated GLM-5.3 routers across discard-and-resample and measure-then-revise.\par}
\end{table}

\FloatBarrier
\subsection{Turn-Level Compute Allocators and Additional Controls}

All three turn-level allocator reimplementations use the same H100 vLLM executor, \texttt{gpt-oss-120b}, and the same $50/65$ retail task partition generated using seed 4321.

Each allocator targets $1.80\times$ the root harness's output-token consumption during calibration, with an admissible search cost of at most $1.84\times$. Held-out rollout seeds are defined as $s \times 100{,}000 + \{1, 2, 3\}$, and evaluations are paired with the corresponding all-low root harness.

No conversation errors or router exceptions occur during these held-out evaluations. All executor computation runs on the self-hosted vLLM server.

\subsubsection{Confidence-Based Turn-Level Routing}

The Confidence-Turn-style allocator estimates uncertainty using the mean negative token log probability of each low-effort generation. When the uncertainty score exceeds a calibrated threshold, the response is regenerated once at medium effort. Both generations contribute to the episode's token consumption.

We evaluate five predefined search quantiles: $q \in \{0.20, 0.35, 0.50, 0.65, 0.80\}$.

Their corresponding search accuracy and token multipliers are $0.327/1.24\times$, $0.313/1.68\times$, $0.393/1.90\times$, $0.367/2.16\times$, and $0.353/2.34\times$.

Under the predefined calibration rule, $q = 0.20$ is selected, achieving a held-out gain of \pmsd{2.6}{2.1}\,pp at $1.31\times$ the baseline output-token consumption.

After observing the initial held-out results, we conduct an additional exploratory search calibration between $q = 0.35$ and $q = 0.50$. The resulting threshold, $q = 0.425$, achieves search accuracy of $0.327$ at $1.79\times$ tokens and a held-out gain of \pmsd{6.3}{3.8}\,pp at $1.81\times$ tokens.

We report this additional result separately because the calibration procedure was motivated by the initial held-out observations.

\subsubsection{Boundary-Aware Turn-Level Routing}

The BAAR-Turn-style allocator fine-tunes \texttt{ModernBERT-base} using $2{,}400$ search episodes with randomly assigned reasoning levels, comprising $31{,}948$ turns.

Ten of the 50 search tasks are reserved for validation. Across three seeds, the success prediction model achieves validation AUC values of $0.595$, $0.579$, and $0.589$. The corresponding token prediction mean absolute errors are $820$, $800$, and $800$ tokens.

The allocator selects reasoning levels by maximizing predicted task success minus a token cost penalty:
\[
\arg\max \left[P(\mathrm{success}) - \lambda \cdot \mathrm{kTokens}\right].
\]

The penalty coefficient $\lambda$ is calibrated toward the $1.80\times$ token target using at most six search evaluations per seed.

One seed consumes $1.74\times$ the baseline tokens during search but increases to $2.04\times$ on held-out tasks. The other two seeds remain eligible and achieve held-out gains of $5.1$ and $7.2$\,pp.

\subsubsection{Reinforcement Learning-Based Compute Allocation}

The CPO-RL-Turn-style allocator uses a $768$--$64$--$3$ multilayer perceptron over frozen \texttt{ModernBERT-base} embeddings. The policy is trained using primal-dual REINFORCE, with a standardized episode-return baseline and a token constraint of $1.80\times$.

Each seed uses $2{,}400$ search episodes, organized into 16 batches of 150 episodes.

The three seeds produce different allocation strategies:
\begin{itemize}[leftmargin=18pt,itemsep=2pt,parsep=0pt,topsep=3pt]
    \item Seed 0 assigns approximately $96\%$ of turns to medium effort and achieves a held-out gain of $11.8$\,pp at $1.56\times$ tokens.
    \item Seed 1 converges toward lower reasoning effort, but the selected policy is the initial untrained policy. It achieves $6.2$\,pp at $1.54\times$ tokens.
    \item Seed 2 selects an almost entirely low-effort policy, achieving $2.6$\,pp at $1.01\times$ tokens.
\end{itemize}

The improvement observed in Seed 2 is consistent with variability between repeated executions of the all-low baseline rather than a substantial change in reasoning allocation.

These results indicate that the learned policies do not consistently discover state-dependent reasoning strategies under the evaluated training configuration.

\subsubsection{Offline Task-Level Routing}

We evaluate whether reasoning effort can be allocated at the task level using only the customer's opening message.

Across nine root rollouts per task, uniform low effort achieves an accuracy of $39.1\%$, while uniform medium effort achieves $48.2\%$. Medium effort outperforms low effort on 26 of the 50 search tasks.

Using these task-level outcomes, we train an L2-regularized logistic regression classifier on TF-IDF features extracted from opening messages.

With regularization parameter $C = 0.1$, the classifier assigns medium effort to 64 of 65 held-out tasks, achieving $9.1$\,pp at $1.63\times$ tokens.

At $C = 1.0$, it assigns medium effort to 41 tasks and achieves $5.5$\,pp at $1.40\times$ tokens. At $C = 10.0$, it assigns medium effort to 39 tasks and achieves $5.5$\,pp at $1.39\times$ tokens.

For comparison, an oracle constructed using observed outcomes on the same 65 evaluation tasks assigns medium effort to 27 tasks, achieving $12.5$\,pp at $1.23\times$ tokens (\cref{tab:offline_router}).

This oracle represents a retrospective reference rather than a deployable task-level routing policy.

\begin{table}[!ht]
\centering
\caption{Offline task-level routing on the 65 held-out tasks (trained on opening user messages across 50 search tasks; 9 root rollouts/tier).}
\label{tab:offline_router}
\vspace{3pt}
\small
\begin{tabular}{@{}lrrrr@{}}
\toprule
Task-Level Policy & Held-Out & Gain vs.\ Low (pp) & Cost & Med.\ Tasks \\
\midrule
All-\texttt{low} (roots) & $0.391$ & $0.0$ & $1.00\times$ & $0/65$ \\
Learned logistic ($C{=}1.0$) & $0.446$ & $5.5$ & $1.40\times$ & $41/65$ \\
Learned logistic ($C{=}0.1$) & $0.482$ & $9.1$ & $1.63\times$ & $64/65$ \\
All-\texttt{medium} (roots) & $0.482$ & $9.1$ & $1.65\times$ & $65/65$ \\
In-sample task oracle & $0.516$ & $12.5$ & $1.23\times$ & $27/65$ \\
\bottomrule
\end{tabular}
\end{table}

\subsubsection{Additional Mechanism Controls}

We conduct additional experiments to determine whether simpler modifications can reproduce the improvements achieved by compute routing.

\textbf{Event routers.} Five Python routers escalate reasoning effort upon write operations, tool errors, conditional user requests, catalog reads, or user turns. None yields substantial held-out gains (the strongest achieves \pmsd{1.0}{1.8}\,pp), as escalating isolated triggers misses multi-step order lookups and unnecessary transfers to a human agent.

\textbf{Human hand-off interception.} Across 345 test episodes, blocking \texttt{transfer\_to\_human\_agents} up to three times reduces accuracy from $35.1\%$ to $31.6\%$. Among the 61 affected tasks, accuracy decreases from $28.4\%$ to $21.3\%$. Inspection of root trajectories shows that 70 of 86 transfers follow explicit customer escalation requests.

\textbf{Fixed information retrieval.} In a separate evaluation using 195 held-out episodes per setting, uniform low effort achieves $39.5\%$ accuracy at $1{,}542$ output tokens per episode, while uniform medium achieves $48.7\%$ at $2{,}806$ tokens. Prefetching customer details achieves $42.1\%$ accuracy at $1.10\times$ the low-effort token consumption.

\textbf{Early-turn reasoning.} Restricting medium reasoning effort to early turns achieves $37.4\%$ accuracy at $1.15\times$ tokens.

Additional search-stage variants that prefetch user or order information achieve accuracies of $30.7\%$ and $28.7\%$, respectively.

These controls do not reproduce the performance of uniform medium effort and suggest that the observed routing improvements cannot be explained solely by these individual modifications.

\subsection{Additional Ablations and Rollout-Scaling Selection Analysis}
\label{app:selection}

\Cref{tab:ablation} reports the individual effects of candidate cost history and discard filtering in GLM-5.2 experiments, and \cref{tab:prose} summarizes the prompt-guideline and guard experiments.

\begin{figure}[!ht]
\centering
\includegraphics[width=0.50\textwidth]{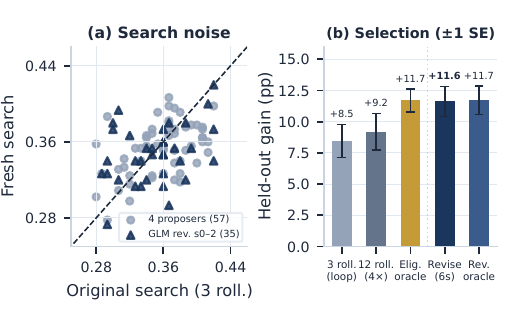}
\vspace{-8pt}
\caption{Selection noise and rollout scaling. (a) Original vs.\ fresh search scores across eligible routers. (b) Held-out gain ($\pm 1\text{ SE}$ across seeds) under $3$ vs.\ $12$ search rollouts alongside eligible oracles.}
\label{fig:selection}
\vspace{-4pt}
\end{figure}

Accurately ranking candidate harnesses remains challenging because of variability in task outcomes. Among budget-eligible routers, search scores obtained from 150 episodes show low correlation with scores obtained from independent search rollouts ($r = 0.14$; \cref{fig:selection}a).

To examine whether additional rollouts improve candidate selection, we reevaluate 57 budget-eligible routers from four proposer settings using nine independent search rollouts per task (\cref{fig:selection}b, \cref{tab:rescore_full}). Combining the original three rollouts with nine additional rollouts (12 rollouts total) increases the mean selected held-out gain from $8.5$ to $9.2$\,pp, compared with an eligible oracle of $11.7$\,pp, closing approximately $22\%$ of the selection gap. Ranking candidates using only the nine additional rollouts achieves $9.5$\,pp.

In a separate comparison, GLM-5.3 measure-then-revise selects routers whose mean performance across six seeds is within $0.1$\,pp of the eligible oracle (\pmsd{11.6}{2.9}\,pp versus \pmsd{11.7}{2.8}\,pp). Reevaluating 35 eligible routers from the first three seeds using three additional search rollouts per task leaves the selected routers unchanged whether ranking uses the original rollouts, the additional rollouts, or all six rollouts (\cref{tab:rescore_full}); the remaining three seeds contain no ties at the highest search score and are not included in this re-scoring experiment. Given that differences in task difficulty account for $69\%$ of episode-level variance, increasing the number of distinct search tasks may be more beneficial than repeatedly evaluating the same tasks, though further experiments are required to establish this.

\begin{table}[!ht]
\centering
\caption{Single-mechanism ablations (GLM-5.2, 3 seeds each).}
\label{tab:ablation}
\vspace{3pt}
\small
\begin{tabular}{@{}lrrrr@{}}
\toprule
Budget Mechanism & Shipped Gain (pp) & Over Cap & $\rho_{\mathrm{start}}$ & Elig.\ Oracle (pp) \\
\midrule
History + discard & \pmsd{9.2}{6.2} & $19/36$ & $34/36$ & \pmsd{12.3}{3.6} \\
History only & \pmsd{4.6}{4.0} & $29/36$ & $34/36$ & \pmsd{7.5}{1.9} \\
Discard only & \pmsd{5.5}{3.1} & $27/36$ & $36/36$ & \pmsd{7.0}{3.3} \\
\bottomrule
\end{tabular}
\end{table}

\begin{table}[!ht]
\centering
\caption{Prose guideline and blocking-guard settings on $\tau$-bench retail (\texttt{gpt-oss-120b}; 3 seeds each). None yields positive transfer.}
\label{tab:prose}
\vspace{3pt}
\small
\begin{tabular}{@{}llrr@{}}
\toprule
Edit Space / Feedback & Split & Shipped $\Delta_{\mathrm{ship}}$ (pp) & $\bar{\tau}_{\mathrm{step}}$ (pp) \\
\midrule
400w prose / failed transcripts & train $\to$ test & \pmsd{-0.6}{0.6} & $-3.6$ \\
150w prose / failed transcripts & train $\to$ test & \pmsd{-3.0}{0.4} & $-1.7$ \\
150w prose / write-action diffs & train $\to$ test & \pmsd{-0.3}{1.1} & $0.0$ \\
150w prose + guard / diffs & train $\to$ test & \pmsd{-0.7}{2.1} & $-4.4$ \\
150w prose / write-action diffs & matched $50/65$ & \pmsd{0.0}{4.9} & $-1.6$ \\
150w prose / contrastive pairs & matched $50/65$ & \pmsd{0.2}{1.6} & $-0.8$ \\
\bottomrule
\end{tabular}
\end{table}

\begin{table}[!ht]
\centering
\caption{Search re-score checks across eligible routers (decimal reward units).}
\label{tab:rescore_full}
\vspace{3pt}
{\small
\textbf{(a) Open-loop discard-and-resample across four proposers ($57$ within-cap routers)}\\[3pt]
\begin{tabular}{@{}lrrrrr@{}}
\toprule
Selection Rule & GLM-5.2 & DeepSeek V4.1 & Gemini 3.8 & Gemini 4 Argon & Mean \\
\midrule
Original 3 rollouts & $0.092$ & $0.079$ & $0.091$ & $0.077$ & $0.085$ \\
9 fresh rollouts only & $0.123$ & $0.101$ & $0.068$ & $0.089$ & $0.095$ \\
Combined 12 rollouts & $0.106$ & $0.104$ & $0.068$ & $0.089$ & $0.092$ \\
Eligible oracle (held-out) & $0.123$ & $0.109$ & $0.125$ & $0.111$ & $0.117$ \\
\bottomrule
\end{tabular}

\vspace{8pt}
\textbf{(b) GLM-5.3 measure-then-revise ($35$ within-cap routers, Seeds 0--2)}\\[3pt]
\begin{tabular}{@{}lrrrr@{}}
\toprule
Selection Rule & Seed 0 & Seed 1 & Seed 2 & Mean $\pm$ SD \\
\midrule
Original 3 rollouts & $0.118$ & $0.067$ & $0.123$ & \pmsd{0.103}{0.031} \\
3 fresh rollouts only & $0.118$ & $0.067$ & $0.123$ & \pmsd{0.103}{0.031} \\
Combined 6 rollouts & $0.118$ & $0.067$ & $0.123$ & \pmsd{0.103}{0.031} \\
Eligible oracle (held-out) & $0.118$ & $0.072$ & $0.123$ & \pmsd{0.104}{0.028} \\
\bottomrule
\end{tabular}
}\\[3pt]
{\raggedright\scriptsize In (a), $57$ within-cap routers are re-scored on $9$ fresh search rollouts ($450$ episodes per router; Gemini 4 Argon pools both recorded re-score batches, while single-batch scoring yields $0.092$ and $0.093$ mean). In (b), $35$ within-cap GLM-5.3 revise routers (Seeds 0--2) are re-scored on $3$ fresh search rollouts ($5{,}250$ episodes total; Seeds 3--5 had no top search-score ties).\par}
\end{table}

\FloatBarrier
\section{Domain Transfer, Additional Executors, and Task Partitions}
\label{app:partitions}

\subsection{Domain Transfer to \texorpdfstring{$\tau$}{tau}-Bench Airline}

We evaluate whether the observed behavior of open-loop router evolution generalizes beyond the retail domain using a disjoint $20/30$ search and held-out task split on $\tau$-bench airline (\cref{tab:full_ledger}). Across three independent seeds with GLM-5.2 as the proposer, 34 of 36 generated routers outperform their root harness on held-out tasks ($\Delta_{\mathrm{all}}^\star = 15.2$\,pp), while 27 of 36 exceed the $2\times$ token budget ($\Delta_{\mathrm{elig}}^\star = 8.1$\,pp). The selected routers achieve a mean held-out improvement of \pmsd{5.2}{5.1}\,pp ($t_{\mathrm{task}} = 1.23$ over $n = 30$ held-out airline tasks). These results reproduce both the positive transfer of executable effort routing and the high budget-violation rate observed under open-loop search in retail.

\subsection{Effect of Reasoning Effort Across Executors}

To examine whether increasing reasoning effort improves performance across different executors, we evaluate three additional models: \texttt{Qwen3.8-27B}, \texttt{Inkling 2 Small}, and \texttt{gpt-oss-20b}. These experiments use uniform reasoning effort without additional harness instructions or compute routers (\cref{tab:exec2}).

\begin{table}[!ht]
\centering
\caption{Uniform effort probes on additional executors (empty harness, no router).}
\label{tab:exec2}
\vspace{3pt}
\resizebox{\textwidth}{!}{%
\begin{tabular}{@{}llrrrrr@{}}
\toprule
Model & Serving Path & Effort & Success (Split) & vs.\ Low (Paired) & Out Tok/Ep (Mult.) & Probe Notes \\
\midrule
\texttt{Qwen/Qwen3.8-27B} & Hosted Tinker & \texttt{low} & $0.569$ ($65$ held-out) & -- & $2{,}730$ ($1.00\times$) & $0$ errors \\
\texttt{Qwen/Qwen3.8-27B} & Hosted Tinker & \texttt{medium} & $0.610$ ($65$ held-out) & $0.041$ ($\text{SE } 0.040$, $t_{\mathrm{task}}=1.02$) & $2{,}869$ ($1.05\times$) & $20/15/30$ up/down/tie \\
\texttt{Qwen/Qwen3.8-27B} & Hosted Tinker & \texttt{xhigh} (\texttt{high}) & $0.579$ ($65$ held-out) & $0.010$ ($\text{SE } 0.033$, $t_{\mathrm{task}}=0.31$) & $3{,}834$ ($1.40\times$) & $14/14/37$ up/down/tie \\
\midrule
\texttt{Inkling 2 Small} & Hosted Tinker & \texttt{low} & $0.626$ ($65$ held-out) & -- & $1{,}647$ ($1.00\times$) & $0/195$ errors \\
\texttt{Inkling 2 Small} & Hosted Tinker & \texttt{medium} & $0.605$ ($65$ held-out) & $-0.021$ ($\text{SE } 0.040$, $t_{\mathrm{task}}=-0.51$) & $2{,}305$ ($1.40\times$) & $16/15/34$ up/down/tie \\
\midrule
\texttt{gpt-oss-20b} & A100 & \texttt{low} & $0.058$ ($115$ test) & -- & $1{,}600$ ($1.00\times$) & $108/115$ always-fail, \texttt{max\_tokens} $4{,}096$ \\
\texttt{gpt-oss-20b} & A100 & \texttt{high} & $0.052$ ($115$ test) & not controlled & $52{,}500$ ($32.4\times$) & $108/115$ always-fail, \texttt{max\_tokens} $8{,}192$ \\
\bottomrule
\end{tabular}%
}\\[2pt]
{\raggedright\scriptsize Hosted Tinker probes ($65$ tasks $\times 3$ rollouts; harmony tokens stripped on \texttt{Qwen3.8-27B}) use hosted simulator calls and are never pooled with H100 vLLM runs. On \texttt{gpt-oss-20b}, \texttt{max\_tokens} changed between \texttt{low} and \texttt{high}.\par}
\end{table}

For \texttt{Qwen3.8-27B}, increasing reasoning effort from low to medium improves accuracy from $56.9\%$ to $61.0\%$, with a $5\%$ increase in output token consumption. However, the improvement is not statistically significant ($t_{\mathrm{task}} = 1.02$). Increasing reasoning effort further to \texttt{xhigh} achieves $57.9\%$ accuracy at $1.40\times$ the low-effort token consumption.

For \texttt{Inkling 2 Small}, increasing reasoning effort from low to medium reduces observed accuracy from $62.6\%$ to $60.5\%$, while increasing output token consumption by $40\%$.

The \texttt{gpt-oss-20b} executor (A100) achieves $5.8\%$ accuracy at low effort and $5.2\%$ at high effort. However, the maximum output token limit differs between the two configurations ($4{,}096$ and $8{,}192$ tokens per call, respectively). This experiment therefore does not isolate the effect of reasoning effort.

The \texttt{Qwen3.8-27B} and \texttt{Inkling 2 Small} experiments use hosted Tinker endpoints and hosted customer-simulator calls. Their results are reported separately from the main experiments conducted on the H100 vLLM setup.

These experiments do not establish consistent improvements from increasing uniform reasoning effort across the additional executors, and we therefore do not perform harness evolution on these models. Together with the retail results on \texttt{gpt-oss-120b}, these observations suggest that the benefits of compute routing depend on the executor's responsiveness to changes in reasoning effort; establishing the applicability of measure-then-revise across additional executors and domains remains an important direction for future work.

\subsection{Task Partitions and Rollout Seeds}

\begin{figure}[!htb]
\centering
\includegraphics[width=0.74\textwidth]{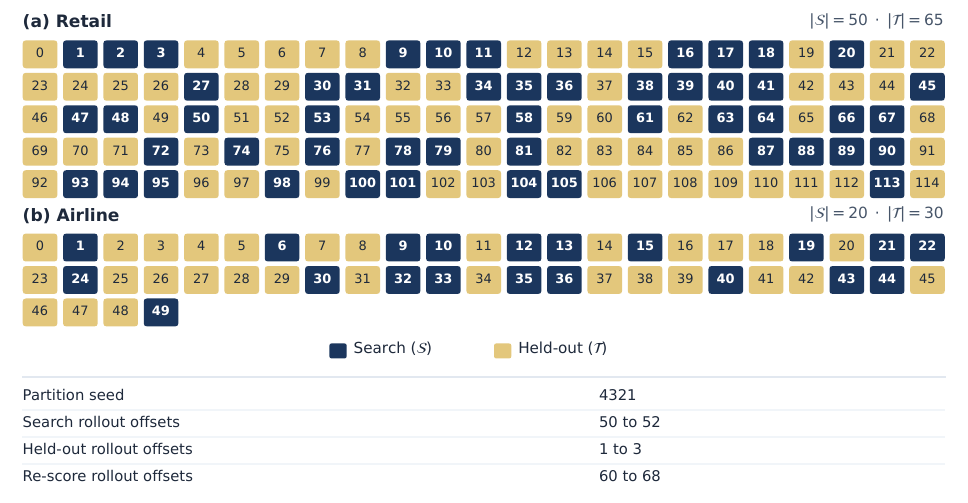}
\vspace{-8pt}
\caption{Exact task-index partitions and rollout seed offsets. Disjoint search ($\mathcal{S}$) and held-out ($\mathcal{T}$) task indices for $\tau$-bench retail ($115$ tasks) and airline ($50$ tasks).}
\label{fig:partitions}
\end{figure}

\Cref{fig:partitions} provides the exact disjoint task partitions generated with partition seed $4321$ for $\tau$-bench retail ($50$ search and $65$ held-out tasks) and airline ($20$ search and $30$ held-out tasks). Search rollouts use seed offsets $50$ to $52$, held-out rollouts use $1$ to $3$, and additional search re-scoring uses $60$ to $68$.

\end{document}